%% file: mopd_domain_generalization_paper.tex
\documentclass{article}

\usepackage{iclr2027_conference,times}
\input{math_commands.tex}

\usepackage{amsmath,amssymb}
\usepackage{algorithm}
\usepackage{algpseudocode}
\usepackage{booktabs}
\usepackage{graphicx}
\usepackage{hyperref}
\usepackage{tabularx}
\usepackage{multirow}
\usepackage{siunitx}
\usepackage{arydshln}
\usepackage{tikz}
\usetikzlibrary{calc}
\usepackage{wrapfig}
\usepackage{url}
\usepackage{xcolor}

\title{Recovering General Capabilities via Uncertainty-Calibrated Multi-Teacher On-Policy Distillation}

\author{%
Ziyuan Liu\textsuperscript{*}\setcounter{footnote}{2}\thanks{Work done during the author's internship at Kuaishou Technology.}\hspace{0.35em},
Jiao Ou\textsuperscript{*\textdagger},
Jian Liang,
Ruiming Tang\textsuperscript{\textdagger}
\& Cheng Luo\\
Kuaishou Technology\\
Beijing, China\\
\texttt{liuziyuan@stu.pku.edu.cn, ojiao1111@gmail.com,}\\
\texttt{\{liangjian03,tangruiming\}@kuaishou.com}\\
\textsuperscript{*}Equal contribution.
\quad
\textsuperscript{\textdagger}Corresponding authors.
}
\iclrfinalcopy

\begin{document}
\maketitle
\pagestyle{plain}

\begin{abstract}
Specializing large language models to vertical domains improves domain-specific behavior but often degrades general capabilities. We study this trade-off in Multi-Teacher On-Policy Distillation (MOPD), where a specialized model learns from domain and general teachers on its own sampled trajectories. Standard MOPD faces two limitations: ordinary on-policy sampling rarely exposes tokens with large positive teacher--student advantages, and advantage sign alone does not establish whether the proposed update direction is reliable. We propose Uncertainty-Calibrated MOPD (UCMOPD), which addresses these limitations through two complementary mechanisms. Golden-Gain Enhancement combines higher-temperature exploration with a standard-temperature anchor and retains trajectories whose positive learning signal matches or exceeds the prompt-specific anchor. Teacher-Endorsement Filtering then uses centered log-likelihood (CLL) to estimate each retained token's plausibility relative to the teacher's uncertainty and probabilistically preserves updates whose directions are supported by that endorsement. Across role-playing and medical-domain specialization, UCMOPD improves the general-capability average over standard MOPD by $4.48\%$ and $7.86\%$, respectively, while maintaining vertical-domain performance. Component ablations and diagnostic analyses support the intended roles of the two mechanisms: exposing and selecting stronger positive signals at the trajectory level and validating update directions through teacher endorsement at the token level.
\end{abstract}

\section{Introduction}

Large language models have demonstrated strong general-purpose capabilities across reasoning, coding, instruction following, and open-ended generation, and are increasingly adapted into vertical-domain systems~\citep{chen2024criticaldomains,lu2024domainadaptation,yang2025specializedllm}. In domains such as role-playing and medical consultation, specialization is essential: the model must internalize domain-specific knowledge and capabilities~\citep{wang2025coser,dou2026baichuan,jin2020disease}. At the same time, practical vertical-domain systems cannot specialize at the expense of general capabilities, because domain tasks often require combining domain knowledge with general reasoning, instruction following, and open-ended communication skills~\citep{liu2024gci,luo2023empirical,dong2024abilities}. After supervised fine-tuning or preference-based optimization on domain data, the model may become better at the target domain while losing part of its general ability, including reasoning, mathematical problem solving, coding, instruction following, and open-ended writing. This domain--general trade-off is especially problematic when the final system is expected to remain a capable general assistant while exhibiting strong domain behavior.

A straightforward remedy is to mix vertical-domain and general-domain data during specialization. Although technically simple, this strategy creates additional practical burdens. Beyond constructing a large, high-quality vertical-domain corpus, developers must invest further effort in collecting and filtering high-quality general data. Even after both sources are available, their mixture ratio must be tuned heuristically: too little general data may fail to preserve general capabilities, whereas too much may weaken domain specialization~\citep{bethune2025scaling}. Maintaining general capability during specialization therefore adds substantial data-construction and mixture-tuning costs. This raises a natural question: can training first focus exclusively on domain capability and then use an open general model to integrate domain and general capabilities into a single model?

Multi-Teacher On-Policy Distillation (MOPD)~\citep{xiao2026mimo,dou2026baichuan,nemotron_cascade2,deepseek_v4_report} offers a viable solution. Training can first focus exclusively on vertical-domain capability to obtain a domain model that serves as the domain-expert teacher, while an open general model serves as the general-expert teacher; MOPD then integrates the capabilities of both experts into a single model. Because vertical-domain models are typically obtained by continuing training from general models, the overall problem can be viewed as restoring general capability after vertical-domain training. A standard setup initializes the student from the vertical-domain model and uses a frozen copy of that model and the original general model as the domain and general teachers, respectively. The student generates responses on vertical-domain and general prompts matched to the two teachers' capability domains, and the corresponding teacher provides supervision, thereby preserving vertical-domain capability while restoring general capability~\citep{camopd_chen2026}.

Effective MOPD generally relies on prompts that adequately cover the capability distributions of the expert teachers~\citep{camopd_chen2026}, yet the original training prompts of open general models are typically unavailable. Publicly available post-training prompts can serve as practical surrogates, but assembling a sufficiently comprehensive set is costly and still unlikely to reproduce the original training distribution. We therefore seek to improve the data efficiency of MOPD, aiming to recover general capabilities effectively from a limited set of publicly available prompts. Given a prompt, the student autoregressively samples a complete rollout trajectory from its own conditional distribution, after which the teacher evaluates each sampled token under the same prefix. The difference between the conditional log-probabilities assigned to that token by the teacher and student defines the distillation advantage,
$A_t=\log \pi_T(y_t\mid q,y_{<t})-\log \pi_S(y_t\mid q,y_{<t})$. When $A_t>0$, the corresponding update increases the student's conditional log-probability of the sampled token; when $A_t<0$, it decreases that conditional log-probability. More importantly, a larger positive advantage induces a stronger positive update, allowing the student to move toward the teacher more rapidly~\citep{minillm}.

\begin{wrapfigure}{r}{0.49\textwidth}
    \vspace{-0.7\baselineskip}
    \centering
    \includegraphics[width=\linewidth]{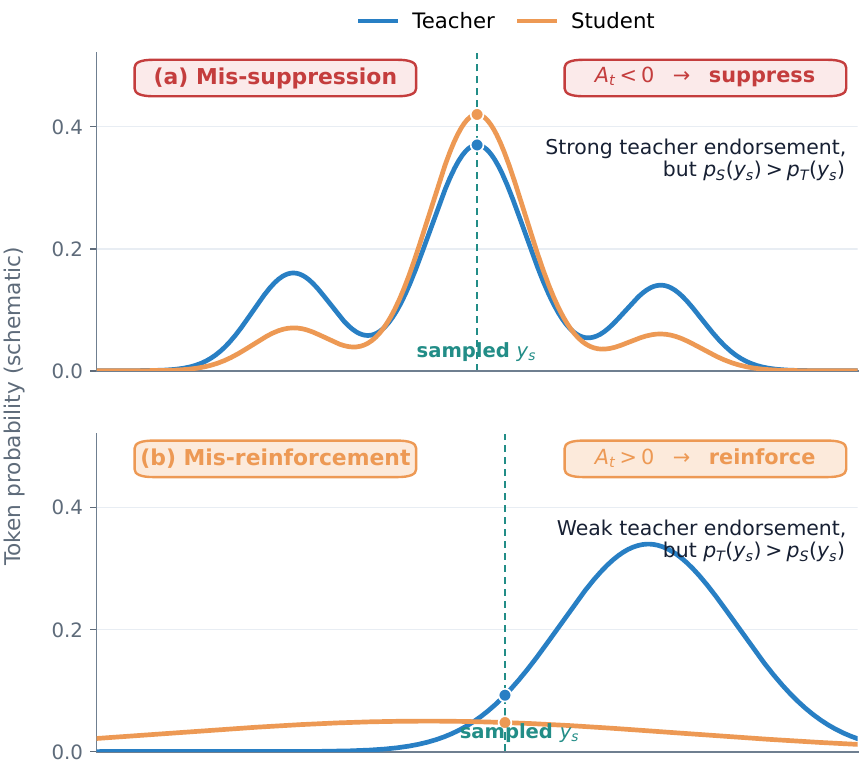}
    \caption{\textbf{Two failure cases of advantage-sign updates.} (a) A strongly teacher-endorsed token can be suppressed when $p_S(y_s)>p_T(y_s)$. (b) A weakly endorsed token can be reinforced when $p_T(y_s)>p_S(y_s)$.}
    \label{fig:negative-advantage-example}
    \vspace{-0.6\baselineskip}
\end{wrapfigure}

Existing MOPD still faces two important problems. (1) Because the student samples each token according to its own conditional distribution, sampled tokens tend to have relatively high student probabilities. Positive advantage requires the teacher probability to exceed the student probability, so large positive advantages are difficult to observe under ordinary sampling, limiting learning from strong positive signals. (2) Using only the sign of advantage to determine whether to increase or decrease the sampled token's probability is not always reliable. As Figure~\ref{fig:negative-advantage-example}(a) illustrates, several plausible continuations may divide the teacher's probability mass, while the student assigns slightly more probability to one of these teacher-endorsed tokens. If that token is sampled, it has $A_t<0$ and is mistakenly suppressed. Conversely, Figure~\ref{fig:negative-advantage-example}(b) shows that a weak student may sample a poorly learned token that the teacher endorses only weakly; if the teacher probability is nevertheless slightly higher than the student's, the token has $A_t>0$ and is mistakenly reinforced. Thus, advantage sign provides only a candidate update direction; its reliability must be assessed independently using the teacher's own endorsement of the current token. Here, \emph{teacher endorsement} refers to the teacher's judgment of the token's plausibility based solely on its own distribution.

The above analysis points to two complementary needs. First, we seek to increase the proportion of signals with large positive advantages in training trajectories; we call these \emph{golden-gain} opportunities. Second, we seek to use teacher endorsement to validate the update direction proposed at each retained token. When the advantage indicates reinforcing a token, stronger teacher endorsement makes that direction more reliable; when it indicates suppressing a token, weaker teacher endorsement makes that direction more reliable.

In this paper, we propose Uncertainty-Calibrated MOPD (UCMOPD) with two complementary mechanisms. Golden-Gain Enhancement uses dual-temperature sampling to generate a standard-temperature anchor and higher-temperature exploration responses, then retains exploration trajectories whose positive learning signal matches or exceeds the anchor. This increases exposure to strong positive advantages without accepting every exploratory trajectory. Inspired by \citet{zhang2026onpolicysft}, Teacher-Endorsement Filtering introduces centered log-likelihood (CLL) to validate the reliability of token-level update directions. For each retained token, CLL assesses endorsement relative to the teacher's own uncertainty: tokens that the teacher considers more plausible provide stronger support for reinforcement, whereas less plausible tokens provide stronger support for suppression. It then probabilistically retains updates whose proposed directions are better supported by the teacher. Experiments on role-playing and medical specialization show that UCMOPD maintains vertical-domain performance. Relative to standard MOPD, it improves the average score across thirteen general-capability benchmarks by $4.48\%$ in role-playing and $7.86\%$ in medical specialization. Component ablations and diagnostic analyses further show that the proposed trajectory- and token-level mechanisms address their intended failure modes.

The contributions of this paper are as follows:
\begin{itemize}
    \item We provide a comprehensive token-level analysis of MOPD signal utilization, showing that advantage sign alone is insufficient for deciding whether teacher feedback should be trusted.
    \item We design UCMOPD with two complementary mechanisms: Golden-Gain Enhancement for discovering and selecting stronger positive signals, and Teacher-Endorsement Filtering for validating token-level update directions.
    \item We validate the method in role-playing and medical-domain specialization, showing that it improves general-capability recovery over vanilla MOPD while preserving or improving vertical-domain performance.
\end{itemize}

\section{Related Work}
\label{sec:related}

\paragraph{On-policy distillation.}
Knowledge distillation transfers predictive structure from teacher to student~\citep{hinton2015distilling}, with sequence-level extensions for autoregressive generation~\citep{kim2016sequence}. OPD instead queries the teacher on student-sampled trajectories to reduce exposure bias, sharing the motivation of interactive imitation learning~\citep{ross2011dagger}. Representative objectives include generalized divergences~\citep{gkd}, reverse-KL policy gradients~\citep{minillm}, and skewed or contrastive losses~\citep{distillm,distillm2}; later work explores relaxed, selective, calibrated, prefix-based, curriculum-style, and related variants~\citep{reopold,selectkd,scope_opd,prefix_opd,paced,skd,adakd,token_selective_dual_kd,eopd_jin2026,yang2026gopd}, as surveyed by \citet{opd_survey_song2026}. Unlike these studies, we ask whether an advantage-proposed update direction agrees with the teacher's own endorsement of the sampled token.

\paragraph{Multi-teacher distillation and capability recovery.}
Multi-teacher distillation aggregates complementary teachers or tasks~\citep{you2017learning,fukuda2017efficient}. Recent LLM systems use on-policy or staged multi-teacher distillation for capability integration, medical adaptation, and reasoning recovery~\citep{xiao2026mimo,dou2026baichuan,nemotron_cascade2,deepseek_v4_report}. Most closely, Counteraction-Aware MOPD selects samples to reduce cross-domain gradient counteraction during capability recovery~\citep{camopd_chen2026}. We instead analyze MOPD at token level and select updates by consistency between their direction and entropy-calibrated teacher endorsement.

\paragraph{Vertical specialization and catastrophic forgetting.}
Vertical specialization can degrade broad abilities, closely relating to catastrophic forgetting~\citep{mccloskey1989catastrophic,kirkpatrick2017overcoming,lopezpaz2017gem,rolnick2019experience,lesort2020continual}. In LLMs, forgetting depends strongly on data composition, task overlap, and general-data injection~\citep{luo2023empirical,dong2024abilities,bethune2025scaling}. Replay or self-synthesized rehearsal~\citep{huang2024self} and general-data injection or capability-integration recipes~\citep{bethune2025scaling,liu2024gci,dong2024abilities} can mitigate it, but require suitable data or carefully curated mixtures. We instead assume that the original general training data are unavailable and recover capability from teacher supervision on student-sampled trajectories.

\section{Method}
\label{sec:method}

In this section, we first review the MOPD objective and distinguish positive learning opportunities from the reliability of advantage-proposed update directions in Section~\ref{sec:background}. We then present Golden-Gain Enhancement, which combines high-temperature exploration with prompt-specific anchor validation to expose and retain stronger positive-advantage signals, in Section~\ref{sec:golden-gain}. Next, we introduce Teacher-Endorsement Filtering for validating token-level update directions in Section~\ref{sec:cll}. Finally, we summarize the complete training procedure in Section~\ref{sec:overall-training}.

\begin{figure*}[t]
    \centering
    \includegraphics[width=\textwidth]{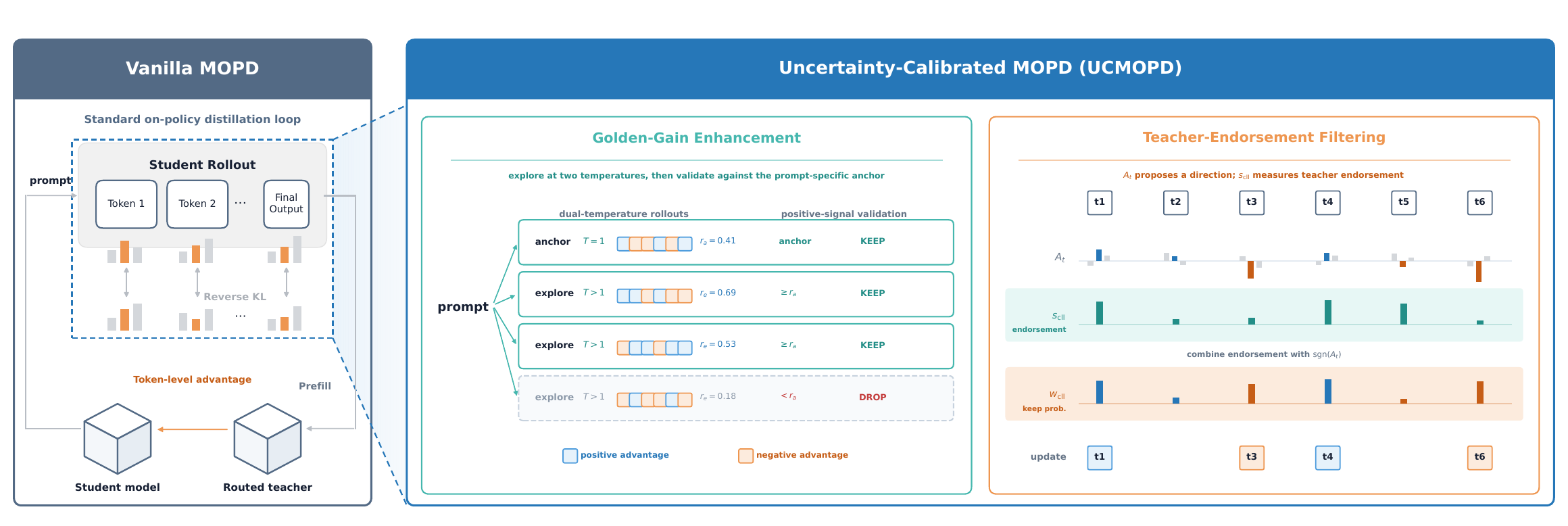}
    \caption{Overview of vanilla MOPD (left) and UCMOPD (right). Golden-Gain Enhancement combines dual-temperature rollouts with prompt-specific trajectory validation. Teacher-Endorsement Filtering distinguishes teacher endorsement $s_{\mathrm{cll}}$ from the update keep probability $w_{\mathrm{cll}}$, which additionally incorporates the advantage sign.}
    \label{fig:method-overview}
\end{figure*}

\subsection{MOPD Background}
\label{sec:background}

We study Multi-Teacher On-Policy Distillation (MOPD) from frozen teacher policies to a student policy $\pi_S$. Each prompt is associated with a domain label $d$ and is routed to the corresponding teacher $\pi_{T(d)}$; for example, role-playing prompts use the domain teacher, while general prompts use the general teacher. This gives a teacher distribution matched to the intended capability of each sampled trajectory without mixing teacher distributions at the token level. Following the on-policy distillation formulation in MOPD~\citep{xiao2026mimo}, the student samples a response $y=(y_1,\ldots,y_T)$ from a rollout policy $\mu_S$, while the training policy is denoted by $\pi_S$. MOPD uses reverse-KL distillation on student-sampled tokens and optimizes the surrogate
\[
    \mathcal L_{\mathrm{MOPD}}(\pi_S)
    =
    -\mathbb E_{q,d,\,y\sim \mu_S}
    \left[
    \frac{1}{T}\sum_{t=1}^{T}
    A^{\mathrm{distill}}_t
    \log \pi_S(y_t\mid q,y_{<t})
    \right],
\]
where the stop-gradient MOPD distillation advantage is
\[
    A^{\mathrm{distill}}_t
    =
    \operatorname{sg}\!\left[
    \log
    \frac{\pi_{T(d)}(y_t\mid q,y_{<t})}
         {\pi_S(y_t\mid q,y_{<t})}
    \right].
\]
Positive values indicate that the routed teacher assigns higher probability than the student to the sampled token, so reinforcing the token moves the student toward the teacher. Negative values indicate that the teacher assigns lower probability than the student, so a naive update suppresses the sampled token. For brevity, we write $A_t$ for $A^{\mathrm{distill}}_t$ and $\pi_T$ for the routed teacher $\pi_{T(d)}$ when the domain label is clear from context.

Token-level MOPD supervision raises two complementary questions. The first is how much the student can learn from a sampled token. For positive advantages, the magnitude of the teacher--student likelihood gap provides a continuous opportunity score: a large positive gap indicates that the token is substantially under-modeled by the student relative to the teacher. We refer to such high-value positive signals as \emph{golden-gain opportunities}. The second question is whether the update direction proposed by the advantage sign is reliable. This requires a teacher-internal endorsement estimate that does not depend on the student probability.

Advantage sign and teacher endorsement therefore play distinct roles. The sign proposes reinforcement or suppression, while endorsement determines whether that direction agrees with the teacher. Table~\ref{tab:direction-endorsement} summarizes the resulting four conceptual cases. The strong/weak endorsement distinction is used only for exposition; Section~\ref{sec:cll} implements the same logic with a continuous keep probability.

\begin{table}[H]
\centering
\small
\setlength{\tabcolsep}{5pt}
\renewcommand{\arraystretch}{1.05}
\caption{Direction--endorsement decomposition of token updates. CLL implements these retention tendencies continuously rather than assigning hard endorsement labels.}
\label{tab:direction-endorsement}
\begin{tabular}{ccl}
\toprule
Advantage direction & Teacher endorsement & Retention tendency \\
\midrule
$A_t>0$ & Strong & More likely to retain reinforcement \\
$A_t>0$ & Weak   & More likely to filter reinforcement \\
$A_t<0$ & Strong & More likely to filter suppression \\
$A_t<0$ & Weak   & More likely to retain suppression \\
\bottomrule
\end{tabular}
\end{table}

\subsection{Golden-Gain Enhancement}
\label{sec:golden-gain}

Standard on-policy sampling concentrates on continuations that the student already assigns high probability. This is appropriate for matching the deployment distribution, but it can repeatedly visit well-modeled regions and provide few opportunities with large positive teacher--student gaps. Increasing the sampling temperature is a natural way to explore less familiar continuations. However, a higher temperature only broadens the candidate pool; it does not guarantee that every resulting trajectory contains stronger positive-advantage signals. We therefore retain a standard-temperature response as a prompt-specific anchor and use it to determine whether each high-temperature exploration trajectory provides a genuine improvement in positive learning signal.

Concretely, for every prompt $q$, we sample one anchor response and $m$ exploration responses,
\[
    y^a \sim \pi_S(\cdot\mid q;T_a),
    \qquad
    y_j^e \sim \pi_S(\cdot\mid q;T_e),
    \quad j=1,\ldots,m,
\]
where $T_a=1.0$ and $T_e>T_a$. The anchor represents the student's ordinary sampling behavior. The exploration branch additionally uses top-$p$ truncation, broadening token coverage while limiting the low-probability tail. To evaluate whether this exploration is effective, we compare each exploratory response against the anchor generated for the same prompt rather than applying a global threshold across prompts of different difficulty and advantage scales.

We score each response with its positive-advantage density,
\[
    r(y)
    =
    \frac{1}{\max(n_+,1)}
    \sum_{t:A_t>0}
    \left(\log \pi_T(y_t\mid q,y_{<t})-\log \pi_S(y_t\mid q,y_{<t})\right),
\]
where $n_+$ is the number of positive-advantage tokens in the response; when $n_+=0$, the score is zero. This score measures the average strength of the response's positive teacher--student gaps. The anchor score $r(y^a)$ defines the prompt-specific threshold: the anchor is always retained, and an exploration response is retained only when
\[
    r(y_j^e) \ge r(y^a).
\]
The comparison keeps exploratory responses whose positive learning signal is at least as strong as the student's ordinary response for the same prompt. In this way, Golden-Gain Enhancement uses higher-temperature sampling to search for golden-gain opportunities and the anchor-based criterion to ensure that only effective exploration contributes to training.

\subsection{Teacher-Endorsement Filtering}
\label{sec:cll}

Our goal is to use teacher endorsement to calibrate the update direction proposed at each retained token. An intuitive approach is to construct the teacher's top-$k$ token set: a student-sampled token inside this set is treated as endorsed, whereas one outside it is treated as unendorsed. However, this decision is sensitive to $k$. A small $k$ may exclude plausible alternatives when the teacher distribution is broad, whereas a large $k$ may include weakly supported tokens when it is sharp.

Inspired by \citet{zhang2026onpolicysft}, we instead use centered log-likelihood (CLL) to measure teacher endorsement; its distribution-discrimination interpretation traces back to Cox's tests of separate families of hypotheses~\citep{cox1961tests}. For a sampled token $x=y_t$, CLL centers its teacher log-likelihood by the teacher entropy,
\[
    c_{\mathrm{cll}}(x)
    =
    \log p_T(x) + H[p_T]
    =
    \log p_T(x)-\mathbb E_{z\sim p_T}[\log p_T(z)].
\]
Because $-H[p_T]=\mathbb E_{z\sim p_T}[\log p_T(z)]$, $c_{\mathrm{cll}}(x)$ compares the sampled token's log-probability with the teacher's probability-weighted mean log-probability over the vocabulary. In particular,
\[
    c_{\mathrm{cll}}(x)\ge 0
    \quad\Longleftrightarrow\quad
    p_T(x)\ge \exp\!\left(-H[p_T]\right).
\]
Thus the reference probability adapts to the shape of the teacher distribution rather than relying on a fixed likelihood or rank threshold. We convert CLL into a bounded teacher-endorsement score,
\[
    s_{\mathrm{cll}}(x)
    =
    \exp\left(
        \min\left(
            \log p_T(x) + H[p_T],
            0
        \right)
    \right),
\]
where $p_T(x)=\pi_T(x\mid q,y_{<t})$ and $H[p_T]$ is the teacher entropy estimate. A token satisfying $p_T(x)\ge \exp(-H[p_T])$ receives maximal endorsement, whereas a token below this entropy-induced typical-probability scale receives proportionally weaker endorsement. This score is independent of advantage sign.

Advantage sign enters only when endorsement is converted into update reliability. Let $d_t=\operatorname{sgn}(A_t)\in\{-1,+1\}$. For every nonzero-advantage token, we define a single keep probability
\[
    w_{\mathrm{cll}}(A_t,x)
    =
    \mathbf 1[A_t>0]s_{\mathrm{cll}}(x)
    +
    \mathbf 1[A_t<0]\bigl(1-s_{\mathrm{cll}}(x)\bigr)
    =
    \frac{1+d_t\bigl(2s_{\mathrm{cll}}(x)-1\bigr)}{2}.
\]
Equivalently, its two directional instances are
\[
\begin{aligned}
    w_{\mathrm{cll}}^{+}(x)
    =
    s_{\mathrm{cll}}(x)
    =
    \exp\!\left(\min\left(\log p_T(x)+H[p_T],0\right)\right),
    \\
    w_{\mathrm{cll}}^{-}(x)
    =
    1-s_{\mathrm{cll}}(x)
    =
    1-\exp\!\left(\min\left(\log p_T(x)+H[p_T],0\right)\right).
\end{aligned}
\]
These are not separate endorsement criteria. CLL always computes the same $s_{\mathrm{cll}}(x)$; the sign merely specifies whether the proposed update increases or decreases the token probability. Strong endorsement agrees with reinforcement and conflicts with suppression, while weak endorsement induces the opposite retention preference.

During training, every nonzero-advantage token receives a Bernoulli mask sampled with probability $w_{\mathrm{cll}}(A_t,y_t)$. Updates whose direction agrees with teacher endorsement therefore remain more often, while endorsement-inconsistent updates are removed more often. No hard endorsement threshold is introduced into the executable rule. We discuss an alternative weighting variant, which multiplies rather than samples the same keep probability, in Appendix~\ref{app:cll-mode}.

The two mechanisms also admit simple local guarantees: Golden-Gain Enhancement cannot retain an exploration trajectory whose positive-advantage density is below the prompt-specific anchor, while Teacher-Endorsement Filtering implements entropy-adaptive direction--endorsement consistency. Appendix~\ref{app:theory} provides the corresponding derivations.

\subsection{Overall Training Procedure}
\label{sec:overall-training}

The resulting training step applies the two mechanisms in sequence. Golden-Gain Enhancement samples anchor and exploratory trajectories and retains those whose positive-advantage density at least matches the prompt-specific anchor baseline. Teacher-Endorsement Filtering then applies CLL to every nonzero-advantage token in the retained trajectories, validating individual update directions against teacher-internal endorsement. This preserves the teacher-guided MOPD objective while making its supervision more selective at both sequence and token levels. Algorithm~\ref{alg:uncertainty-calibrated-mopd} summarizes the complete procedure.

\begin{algorithm}[t]
\caption{UCMOPD}
\label{alg:uncertainty-calibrated-mopd}
\small
\begin{algorithmic}[1]
\Require Student policy $\pi_S$, teacher policies $\{\pi_T\}$, prompt batch $\mathcal{Q}$, anchor temperature $T_a=1.0$, exploration temperature $T_e$
\Ensure Updated student policy $\pi_S$
\For{each prompt $q \in \mathcal{Q}$}
    \State Sample anchor response $y^{a}\sim \pi_S(\cdot\mid q;T_a)$
    \State Sample exploration responses $\{y^{e}_j\}_{j=1}^{m}\sim \pi_S(\cdot\mid q;T_e)$
    \State $\mathcal{Y}_q \gets \{y^{a}\}\cup\{y^{e}_j\}_{j=1}^{m}$
    \For{each response $y\in\mathcal{Y}_q$}
        \State Compute student and teacher log probabilities for each token
        \State Compute token advantages $A_t=\log \pi_T(y_t\mid q,y_{<t})-\log \pi_S(y_t\mid q,y_{<t})$
        \State Compute positive-advantage density $r(y)$ over tokens with $A_t>0$
    \EndFor
    \State Keep $y^a$ and keep each exploration response $y^e_j$ only if $r(y^e_j)\ge r(y^a)$
    \For{each retained response $y$}
        \For{each token position $t$ with $A_t\ne0$}
            \State Estimate teacher entropy $H_T(t)$ at position $t$
            \State $s_t\gets \exp(\min(\log \pi_T(y_t\mid q,y_{<t})+H_T(t),0))$
            \State $w_t\gets \mathbf 1[A_t>0]s_t+\mathbf 1[A_t<0](1-s_t)$
            \State Keep token update $t$ with probability $w_t$
        \EndFor
    \EndFor
\EndFor
\State Update $\pi_S$ with the MOPD loss on retained responses and retained tokens
\end{algorithmic}
\end{algorithm}

\section{Experiments}
\label{sec:experiments}

\subsection{Experimental Setup}

We evaluate domain specialization recovery in two vertical settings: role-playing and medical-domain adaptation. In each setting, the student is initialized from the supervised domain-specialized model, and MOPD uses a domain teacher and a general teacher to provide token-level on-policy supervision. We compare against the original general model, the supervised fine-tuned (SFT) domain model, vanilla MOPD, and three additional distillation baselines: SelecTKD~\citep{selectkd}, ReOPOLD~\citep{reopold}, and CaMOPD~\citep{camopd_chen2026}. Our dual-temperature rollout uses one anchor response at temperature $1.0$ in both domains. The role-playing setting uses seven exploration responses at temperature $1.5$, while the medical setting uses three exploration responses at temperature $1.5$. Appendix~\ref{app:full-setup} provides the complete checkpoints, teacher--student scale relationship, routing rules, prompt sources, precision, and optimization configuration.

We report both general capabilities and vertical-domain performance. General capability is measured by MMLU-Redux~\citep{gema2024done}, MMLU-Pro~\citep{wang2024mmlu}, GPQA-Diamond~\citep{rein2023gpqa}, AIME25 and HMMT25~\citep{balunovic2025matharena}, ZebraLogic~\citep{lin2025zebralogic}, LiveCodeBench v5 and v6~\citep{jain2024livecodebench}, IF-Eval~\citep{zhou2023ifeval}, IFBench~\citep{pyatkin2025generalizing}, WritingBench~\citep{wu2025writingbench}, Arena-Hard v2~\citep{li2024arenahard}, and LiveBench 1125~\citep{white2025livebench}. Role-playing ability is measured by storyline consistency, anthropomorphism, character fidelity, and storyline quality following the CoSER evaluation setting~\citep{wang2025coser}. Medical-domain ability is measured by MedQA-USMLE~\citep{jin2020disease}, MedXpertQA Text~\citep{zuo2025medxpertqa}, and PubMedQA~\citep{jin2019pubmedqa}. Unless otherwise stated, higher is better for all metrics.

\subsection{Main Results}

\begin{figure}[H]
\centering
\begin{minipage}[t]{0.485\linewidth}
    \vspace{0pt}
    \centering
    \includegraphics[width=\linewidth]{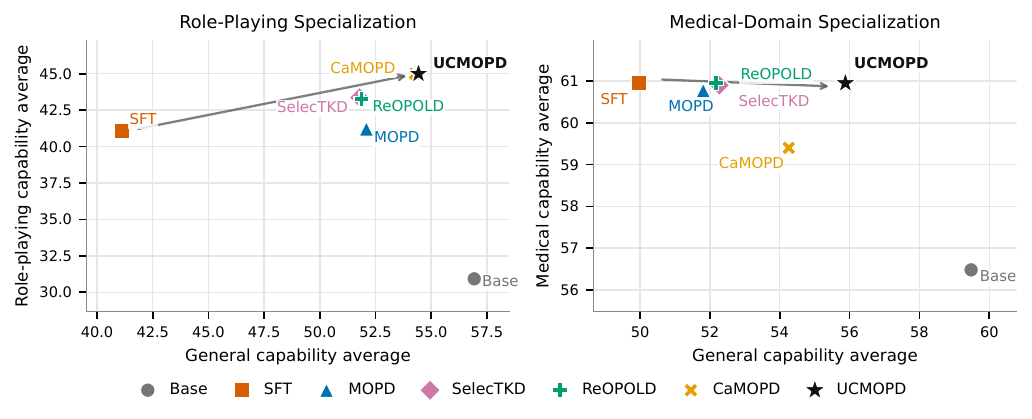}
    \caption{Domain--general trade-off in role-playing and medical specialization. UCMOPD improves general recovery while preserving vertical-domain ability.}
    \label{fig:capability-tradeoff}
\end{minipage}
\hfill
\begin{minipage}[t]{0.485\linewidth}
    \vspace{0pt}
    \centering
    \includegraphics[width=\linewidth]{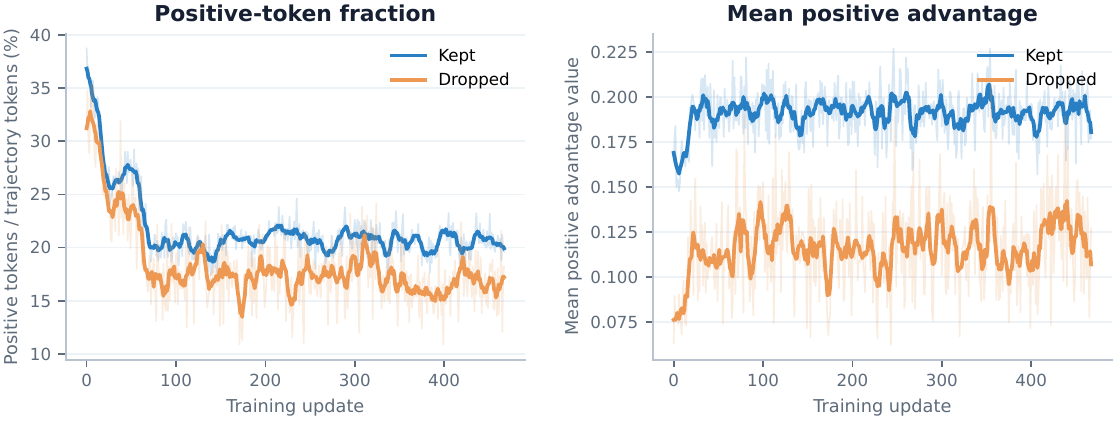}
    \caption{Training-update diagnostics of Golden-Gain Enhancement before Teacher-Endorsement Filtering. Retained trajectories exhibit a higher positive-token fraction and a higher mean positive advantage.}
    \label{fig:analysis-diagnostics}
\end{minipage}
\end{figure}

Table~\ref{tab:main-results} and Figure~\ref{fig:capability-tradeoff} summarize the role-playing results. SFT substantially improves role-playing ability, raising the vertical average from $30.92$ to $41.08$, but it causes a large drop in the general average from $56.93$ to $41.11$. Vanilla MOPD recovers much of the general ability ($52.09$) but barely improves the role-playing average over SFT ($41.22$). UCMOPD obtains a strong trade-off among the compared methods: it reaches the best general average of $54.43$, slightly above CaMOPD's $54.27$, and the best role-playing average ($45.00$, tied with CaMOPD). Compared with vanilla MOPD, UCMOPD improves the general average by $+2.33$ points and the role-playing average by $+3.78$ points.

The medical-domain results show a complementary pattern. Medical SFT improves the medical average from $56.48$ to $60.96$, but reduces the general average from $59.48$ to $49.97$. Vanilla MOPD recovers only part of the general capability ($51.81$). In contrast, UCMOPD reaches a general average of $55.88$, improving over vanilla MOPD by $+4.07$ points while maintaining a medical average of $60.95$, essentially matching SFT ($60.96$) and exceeding CaMOPD ($59.40$). This indicates that the proposed selection mechanism is especially helpful when the domain-specialized model suffers a large general-capability drop.

\begin{table}[H]
\centering
\caption{Main specialization results for role-playing (left) and medical-domain (right) specialization. Base and SFT denote the pre-specialization and supervised domain-specialized models; the remaining columns are general-capability recovery methods. Bold marks the best recovery-method result per metric. ``Gen. Avg.'' averages general benchmarks, while ``Vertical Avg.'' averages the listed vertical benchmarks.}
\label{tab:main-results}
\begin{tikzpicture}[baseline]
\node[inner sep=0pt,outer sep=0pt] (main-results-table) {\resizebox{\linewidth}{!}{%
\begin{tabular}{l*{7}{S[table-format=2.2]}*{7}{S[table-format=2.2]}}
\toprule
\multirow{2}{*}{Metric} & \multicolumn{7}{c}{\textit{Role-playing specialization}} & \multicolumn{7}{c}{\textit{Medical-domain specialization}} \\
\cmidrule(lr){2-8}\cmidrule(lr){9-15}
& \multicolumn{1}{c}{Base} & \multicolumn{1}{c}{SFT} & \multicolumn{1}{c}{MOPD} & \multicolumn{1}{c}{SelecTKD} & \multicolumn{1}{c}{ReOPOLD} & \multicolumn{1}{c}{CaMOPD} & \multicolumn{1}{c}{UCMOPD}
& \multicolumn{1}{c}{Base} & \multicolumn{1}{c}{SFT} & \multicolumn{1}{c}{MOPD} & \multicolumn{1}{c}{SelecTKD} & \multicolumn{1}{c}{ReOPOLD} & \multicolumn{1}{c}{CaMOPD} & \multicolumn{1}{c}{UCMOPD} \\
\midrule
MMLU-Redux & 82.39 & 75.05 & 78.86 & 78.60 & 78.65 & 79.07 & \textbf{80.56} & 84.67 & 85.77 & 86.21 & \textbf{86.30} & 86.21 & 86.21 & 86.11 \\
MMLU-Pro & 68.33 & 54.45 & 62.20 & 61.60 & 62.36 & 64.31 & \textbf{65.72} & 73.08 & 74.80 & 74.53 & \textbf{74.67} & 74.14 & 74.53 & 74.04 \\
GPQA-Diamond & 61.49 & 54.29 & 59.60 & 60.98 & 63.51 & 62.63 & \textbf{63.64} & 62.50 & 62.37 & 60.10 & 58.46 & 61.49 & \textbf{61.99} & 61.74 \\
AIME25 & 45.21 & 40.62 & \textbf{47.29} & 45.00 & 43.12 & 45.21 & 43.54 & 65.00 & 41.67 & 54.37 & 53.75 & 52.08 & 54.37 & \textbf{56.25} \\
ZebraLogic & 79.50 & 31.50 & 71.10 & 70.60 & 71.40 & \textbf{76.30} & 76.20 & 84.80 & 71.60 & 72.60 & 72.10 & 73.20 & 72.60 & \textbf{73.70} \\
HMMT25 & 33.12 & 26.67 & 31.04 & 31.04 & 31.25 & 31.04 & \textbf{32.08} & 45.00 & 28.33 & 33.96 & 33.54 & 32.92 & 34.58 & \textbf{35.62} \\
LiveCodeBench v5 & 34.76 & 25.09 & 35.12 & 34.41 & 32.26 & 37.28 & \textbf{39.78} & 58.42 & 48.75 & 51.25 & 54.48 & 53.76 & \textbf{56.63} & 54.84 \\
LiveCodeBench v6 & 20.57 & 16.00 & \textbf{25.14} & 24.00 & 22.86 & 24.00 & \textbf{25.14} & 31.43 & 27.43 & 24.00 & \textbf{29.71} & 29.14 & \textbf{29.71} & 27.43 \\
IF-Eval & 83.18 & 67.65 & 78.00 & 78.93 & 79.67 & 80.41 & \textbf{80.96} & 84.66 & 48.98 & 48.06 & 49.17 & 48.61 & 59.89 & \textbf{71.53} \\
IFBench & 30.95 & 23.81 & 26.53 & 26.53 & 26.19 & 26.87 & \textbf{27.55} & 21.09 & 22.45 & \textbf{22.11} & 20.41 & 21.09 & \textbf{22.11} & 21.09 \\
WritingBench & 83.80 & 69.58 & 80.30 & 80.53 & 80.39 & 81.64 & \textbf{81.71} & 79.87 & 79.42 & 79.62 & 79.78 & \textbf{79.84} & 79.62 & 79.73 \\
Arena-Hard v2 & 51.15 & 6.95 & 22.60 & 22.00 & 23.70 & \textbf{33.25} & 29.35 & 30.65 & 23.65 & 23.90 & 25.40 & 25.15 & 24.10 & \textbf{26.05} \\
LiveBench 1125 & 65.60 & 42.80 & 59.40 & 58.90 & 59.00 & \textbf{63.50} & 61.30 & 52.10 & 34.40 & 42.80 & 41.80 & 40.70 & 49.00 & \textbf{58.30} \\
Storyline Consistency & 34.74 & 42.27 & 43.57 & 43.87 & 45.42 & \textbf{47.81} & 47.39 & \multicolumn{7}{c}{--} \\
Anthropomorphism & 23.09 & 33.42 & 34.91 & 32.28 & 33.35 & 33.96 & \textbf{35.96} & \multicolumn{7}{c}{--} \\
Character Fidelity & 23.89 & 38.75 & 41.45 & 39.46 & 41.53 & \textbf{44.65} & 42.98 & \multicolumn{7}{c}{--} \\
Storyline Quality & 41.98 & 49.89 & 51.50 & 51.54 & 52.84 & 53.47 & \textbf{53.67} & \multicolumn{7}{c}{--} \\
MedQA-USMLE & \multicolumn{7}{c}{--} & 79.03 & 86.17 & 86.02 & 85.78 & 85.94 & 85.78 & \textbf{86.10} \\
MedXpertQA Text & \multicolumn{7}{c}{--} & 17.22 & 22.90 & 23.55 & 23.51 & \textbf{24.12} & 23.63 & 23.06 \\
PubMedQA & \multicolumn{7}{c}{--} & 73.19 & 73.81 & 72.77 & 73.41 & 54.07 & 68.79 & \textbf{73.69} \\
\midrule
Gen. Avg. & 56.93 & 41.11 & 52.09 & 51.78 & 51.87 & 54.27 & \textbf{54.43} & 59.48 & 49.97 & 51.81 & 52.27 & 52.18 & 54.26 & \textbf{55.88} \\
Vertical Avg. & 30.92 & 41.08 & 41.22 & 43.36 & 43.28 & \textbf{45.00} & \textbf{45.00} & 56.48 & 60.96 & 60.78 & 60.90 & 54.71 & 59.40 & \textbf{60.95} \\
\bottomrule
\end{tabular}%
}};
\path (main-results-table.north west) -- (main-results-table.north east)
  coordinate[pos=0.565] (main-results-sep-top);
\path (main-results-table.south west) -- (main-results-table.south east)
  coordinate[pos=0.565] (main-results-sep-bottom);
\draw[dash pattern=on 4pt off 4pt,line width=0.4pt]
  ([yshift=-18pt]main-results-sep-top) --
  ([yshift=1pt]main-results-sep-bottom);
\end{tikzpicture}
\end{table}

\subsection{Ablations}

Unless otherwise specified, ablations and diagnostics use the role-playing specialization setting. Table~\ref{tab:component-ablation} starts from full UCMOPD and cumulatively removes positive-advantage-density filtering, dual-temperature sampling, and Teacher-Endorsement Filtering. Removing either Golden-Gain operation weakens the overall trade-off, while removing Teacher-Endorsement Filtering further lowers both averages.

\begin{table}[H]
\centering
\footnotesize
\setlength{\tabcolsep}{5pt}
\caption{Cumulative component removal from UCMOPD on role-playing specialization.}
\label{tab:component-ablation}
\resizebox{\linewidth}{!}{%
\begin{tabular}{l*{6}{S[table-format=2.2]}}
\toprule
Method & \multicolumn{1}{c}{Gen. Avg.} & \multicolumn{1}{c}{Vertical Avg.} & \multicolumn{1}{c}{IF-Eval} & \multicolumn{1}{c}{GPQA} & \multicolumn{1}{c}{Zebra} & \multicolumn{1}{c}{Arena-Hard v2} \\
\midrule
UCMOPD & \textbf{54.43} & \textbf{45.00} & \textbf{80.96} & \textbf{63.64} & \textbf{76.20} & \textbf{29.35} \\
$-$ Positive-advantage-density filtering & 53.32 & 43.24 & 80.04 & 63.13 & 74.30 & 28.45 \\
$-$ Dual-temperature sampling & 52.17 & 43.36 & 77.63 & 62.63 & 72.00 & 23.70 \\
$-$ Teacher-Endorsement Filtering $=$ MOPD & 52.09 & 41.22 & 78.00 & 59.60 & 71.10 & 22.60 \\
\bottomrule
\end{tabular}
}
\end{table}

\begin{table}[t]
\centering
\footnotesize
\setlength{\tabcolsep}{4pt}
\caption{Token-endorsement criterion comparison. Both criteria retain updates whose direction agrees with estimated teacher endorsement; CLL replaces hard top-$k$ membership with a continuous entropy-centered score.}
\label{tab:cll-vs-topk}
\resizebox{\linewidth}{!}{%
\begin{tabular}{l*{6}{S[table-format=2.2]}}
\toprule
Criterion & \multicolumn{1}{c}{Gen. Avg.} & \multicolumn{1}{c}{Vertical Avg.} & \multicolumn{1}{c}{AIME25} & \multicolumn{1}{c}{GPQA} & \multicolumn{1}{c}{Zebra} & \multicolumn{1}{c}{Arena-Hard v2} \\
\midrule
Hard top-$k$ endorsement ($k=3$) & 51.95 & 42.31 & 43.33 & 60.73 & \textbf{74.40} & 23.20 \\
Hard top-$k$ endorsement ($k=1$) & 52.01 & 41.32 & 42.71 & \textbf{63.26} & 71.90 & 22.65 \\
CLL sample mask & \textbf{52.17} & \textbf{43.36} & \textbf{44.58} & 62.63 & 72.00 & \textbf{23.70} \\
\bottomrule
\end{tabular}%
}
\end{table}

To verify that the improvement is not simply due to more responses per prompt, we compare against a rollout-budget-matched vanilla MOPD baseline. Both methods use eight responses per prompt, and UCMOPD achieves higher general and role-playing averages; Appendix Table~\ref{tab:rollout-budget-control} reports the full results.

We use an exploration temperature of $T=1.5$ because it yields the strongest role-playing average among the tested temperatures; the full sensitivity study is in Appendix~\ref{app:temperature-ablation}.

\subsection{Analysis of the Proposed Mechanisms}
\label{sec:analysis}

\paragraph{Golden-Gain Enhancement.}
Ordinary on-policy sampling rarely exposes large positive teacher--student gaps. Golden-Gain Enhancement broadens the candidate pool through dual-temperature sampling and retains exploration trajectories only when their positive-advantage density matches or exceeds the prompt-specific anchor. In Table~\ref{tab:component-ablation}, removing trajectory filtering reduces the general average from $54.43$ to $53.32$ and the vertical average from $45.00$ to $43.24$; further removing dual-temperature sampling reduces the general average to $52.17$. Appendix Table~\ref{tab:rollout-budget-control} rules out rollout count as the explanation. Figure~\ref{fig:analysis-diagnostics} further shows that retained trajectories consistently contain stronger positive signals: their aggregate positive-token fraction and mean positive advantage are $0.2168$ and $0.1911$, compared with $0.1850$ and $0.1158$ for rejected trajectories (Appendix Table~\ref{tab:positive-advantage-density-diagnostics}). Although only positive-advantage magnitude directly determines selection, the higher positive-token fraction indicates an empirical enrichment: trajectories with stronger positive signals also tend to contain such signals more frequently. Thus, Golden-Gain Enhancement exposes more candidates and concentrates training on stronger positive learning opportunities.

\paragraph{Teacher-Endorsement Filtering.}
Advantage sign compares teacher and student probabilities but does not measure the teacher's own endorsement, so a strongly endorsed token may be suppressed or a weakly endorsed token reinforced, as Figure~\ref{fig:negative-advantage-example} illustrates. Teacher-Endorsement Filtering therefore uses CLL to evaluate each retained token according to direction--endorsement consistency. Removing this mechanism from the CLL-only configuration reduces the general average from $52.17$ to $52.09$ and the vertical average from $43.36$ to $41.22$ (Table~\ref{tab:component-ablation}). Table~\ref{tab:cll-vs-topk} also shows that entropy-centered CLL gives the best general and vertical averages among the tested endorsement criteria, outperforming fixed top-$k$ endorsement, which ignores distribution shape. Together, the results confirm the intended division of labor: Golden-Gain Enhancement discovers stronger positive opportunities, while Teacher-Endorsement Filtering validates the proposed token-level update directions. Additional direction--endorsement diagnostics and realized CLL keep rates are reported in Appendix~\ref{app:teacher-endorsement-diagnostics}.

\section{Conclusion}
\label{sec:conclusion}

We study general-capability recovery after vertical-domain specialization as a signal-selection problem in Multi-Teacher On-Policy Distillation. UCMOPD uses Golden-Gain Enhancement to expose and select trajectories with stronger positive learning signals and Teacher-Endorsement Filtering to validate token-level update directions. Across role-playing and medical specialization, UCMOPD improves the general-capability average over standard MOPD by $4.48\%$ and $7.86\%$, respectively, while maintaining vertical-domain performance. Ablations and diagnostic analyses further show that these gains do not merely arise from a larger rollout budget and that the two mechanisms address their intended failure modes. These results support uncertainty-aware signal discovery and validation as an effective approach to preserving general capabilities during domain specialization.

\subsection*{AI use statement}

In this work, we used generative AI tools for assisting with translation, supporting qualitative and thematic data analysis, interpreting results. We have not used generative AI tools for generating synthetic data sets, helping to develop theoretical models or conceptual frameworks, assisting in the writing of proofs, proposing or refining hypotheses, designing or providing feedback on research methodology or experiments, implementing methods, cleaning and reformatting dataset. Formulating mathematical claims, providing critical ingredients for proving mathematical claims are not applicable to this work. Additionally, we used generative AI tools for creating or editing software code, creating artifacts, sourcing/searching for information, editing a research paper to improve readability. We have reviewed all AI-assisted work. AI-assisted text and analyses were checked against the source materials and experimental results, and AI-assisted code was manually reviewed and tested for correctness by the authors. We take responsibility for the final content of this work, including text, claims or artifacts produced with the aid of generative AI.

\subsection*{Ethics Statement}

This work uses publicly available language-model checkpoints and benchmark datasets. It does not involve human-subject experiments, collection of personal information, or interaction with identifiable individuals. The role-playing and medical-domain evaluations use existing benchmark prompts and follow their intended evaluation protocols. As with other language-model research, generated outputs may contain factual errors or undesirable content; they should not be treated as professional medical advice or as a substitute for human review.

\subsection*{Reproducibility Statement}

We provide the algorithm, training objective, filtering rules, model roles, prompt-mixture construction, rollout temperatures, precision settings, and optimization configuration in the main paper and Appendix~\ref{app:full-setup}. The benchmark definitions and citations, complete per-benchmark results, ablations, and diagnostic statistics are also reported in the paper or its appendices. These details are intended to make the reported comparisons and training procedure reproducible using the cited public checkpoints and datasets.

\bibliographystyle{iclr2027_conference}
\bibliography{references}

\appendix

\section{Detailed Experimental Setup}
\label{app:full-setup}

\subsection{Models and Systems}

For role-playing, a CoSER-fine-tuned Qwen3-4B-Instruct-2507 checkpoint~\citep{yang2025qwen3technicalreport,wang2025coser} initializes the student and serves as the frozen domain teacher; the original checkpoint serves as the general teacher. Medical adaptation uses II-Medical-7B-Preview~\citep{ii_medical_7b_preview} for student initialization and the domain teacher, with Qwen3-8B as the general teacher. Table~\ref{tab:checkpoint-setup} summarizes these model roles.

Both settings therefore use \emph{homogeneous, same-scale} MOPD: the student and both teachers share the same Qwen3 architecture and parameter count. The teachers differ in specialization rather than scale, and only the student parameters are updated.

\begin{table}[H]
\centering
\small
\setlength{\tabcolsep}{4pt}
\renewcommand{\arraystretch}{1.08}
\caption{Model checkpoints and precision. ``Same as student initialization'' means that the domain teacher is a frozen copy of the SFT checkpoint from which the trainable student is initialized.}
\label{tab:checkpoint-setup}
\begin{tabularx}{\textwidth}{@{}l l >{\raggedright\arraybackslash}X c c c@{}}
\toprule
Setting & Role & Checkpoint & Size & Training dtype & Inference dtype \\
\midrule
Role-playing & Student & Our CoSER SFT of Qwen3-4B-Instruct-2507 & 4B & BF16 & BF16 \\
Role-playing & Domain teacher & Same as student initialization (frozen) & 4B & -- & BF16 \\
Role-playing & General teacher & Qwen3-4B-Instruct-2507 (frozen) & 4B & -- & BF16 \\
\midrule
Medical & Student & II-Medical-7B-Preview & 8B & BF16 & BF16 \\
Medical & Domain teacher & Same as student initialization (frozen) & 8B & -- & BF16 \\
Medical & General teacher & Qwen3-8B (frozen) & 8B & -- & BF16 \\
\bottomrule
\end{tabularx}
\end{table}

Student optimization uses Megatron-Core with BF16 parameters, while student rollout generation and both teachers' log-probability inference use asynchronous SGLang with BF16 weights; no model is quantized. Actor training uses pipeline parallelism of 2 for role-playing and 4 for the larger medical model. Student and teacher rollout inference use tensor parallelism of 1. Both settings run on 4 nodes with 8 NVIDIA H800 GPUs per node.

\subsection{Data Construction and Routing}

Each recovery corpus concatenates a vertical-domain prompt set with a general prompt set and is globally shuffled before training. The role-playing corpus contains 10,000 CoSER prompts and 10,000 general prompts. The medical corpus contains 9,926 medical prompts and the same 10,000 general prompts. The two mixtures are therefore exactly $1{:}1$ and approximately $1{:}1$, respectively.

\paragraph{Role-playing data.}
We rank the CoSER conversations~\citep{wang2025coser} by their number of dialogue turns and reserve the 10,000 longest conversations as unseen role-playing prompts for MOPD. For each reserved conversation, the first $n-1$ turns form the prompt and context, while the final continuation is generated on-policy by the current student. These reserved conversations are never used for role-playing SFT. The remaining approximately 300,000 conversations are used to fine-tune Qwen3-4B-Instruct-2507 and obtain the role-playing initialization and domain teacher.

\paragraph{General recovery data.}
We draw the general prompts from the chat, code, math, and STEM subsets of the Nemotron Post-Training Dataset v1~\citep{bercovich2025llamanemotron}. We sample 10,000 prompts while preserving the relative proportions of these four selected subsets. MOPD uses only the prompt and conversational context; reference responses from the source dataset are not included in the distillation loss.

\paragraph{Medical data.}
We construct the medical prompt pool from four public sources: 4,000 prompts from MedMCQA~\citep{pal2022medmcqa}, 3,000 from MedReason~\citep{wu2025medreason}, 2,000 English prompts from Medical-R1-Distill-Data~\citep{chen2024huatuogpto1}, and 1,000 from \mbox{m23k-tokenized} \mbox{\citep{huang2025m1}}. After preprocessing, 9,926 valid prompts remain. This stream is prompt-only: gold answers and explanations are not provided during MOPD training. The current student generates on-policy responses, and the medical teacher supplies token-level log-probability feedback on those responses.

\paragraph{Overlap control and teacher routing.}
We verify that none of the role-playing, general, or medical training prompts overlaps with any instance in the evaluation sets used in this paper. Teacher routing is deterministic and defined at the prompt level. Each example carries a source label that is preserved through rollout generation. Role-playing labels \texttt{coser} and \texttt{general} route to the CoSER SFT teacher and Qwen3-4B-Instruct-2507 teacher, respectively; medical labels \texttt{medical} and \texttt{general} route to the medical SFT teacher and Qwen3-8B teacher, respectively. All anchor and exploration responses sampled from one prompt inherit the same label and are scored by the same routed teacher at every token. We do not average, interpolate, or otherwise mix teacher distributions within a trajectory.

\subsection{Training Configuration}

Table~\ref{tab:training-setup} reports the optimization and sampling configuration. A ``prompt batch'' counts distinct prompts before rollout replication. The role-playing setting samples eight responses per prompt (one anchor and seven exploration responses), whereas the medical setting samples four (one anchor and three exploration responses); both therefore process 1,024 candidate responses per outer update before positive-advantage-density filtering. The anchor uses temperature $1.0$ and top-$p=1.0$, and exploration uses temperature $1.5$ and top-$p=0.9$. To reduce memory and communication overhead, the teacher entropy used by CLL is estimated from the teacher's top-50 log-probabilities, $\widehat{H}_T(t)=-\sum_{i=1}^{50}p_T^{(i)}(t)\log p_T^{(i)}(t)$, rather than from the full vocabulary; this top-$k$ approximation avoids materializing and transferring the full distribution. We train for three passes over each shuffled prompt mixture. Because the dataloader drops the last incomplete batch, this gives $\lfloor 20000/128\rfloor\times3=468$ outer updates for role-playing and $\lfloor19926/256\rfloor\times3=231$ for medical adaptation.

\begin{table}[H]
\centering
\small
\setlength{\tabcolsep}{4pt}
\renewcommand{\arraystretch}{1.05}
\caption{Training and rollout configuration. Batch sizes are global prompt counts before response replication.}
\label{tab:training-setup}
\begin{tabular}{lrr}
\toprule
Configuration & \multicolumn{1}{c}{Role-playing} & \multicolumn{1}{c}{Medical} \\
\midrule
Domain prompts & 10,000 & 9,926 \\
General prompts & 10,000 & 10,000 \\
Global prompt batch size & 128 & 256 \\
Responses per prompt & 8 ($1{+}7$) & 4 ($1{+}3$) \\
Candidate responses/update & 1,024 & 1,024 \\
Training epochs & 3 & 3 \\
Outer training updates & 468 & 231 \\
Actor updates per batch & 1 & 1 \\
Learning rate & $2\times10^{-6}$ & $2\times10^{-6}$ \\
Gradient clipping & 1.0 & 1.0 \\
Prompt/response limit & 4,096/32,768 & 4,096/32,768 \\
\bottomrule
\end{tabular}
\end{table}

The two settings use different thinking configurations. Role-playing uses the non-thinking configuration of Qwen3-4B-Instruct-2507, so the student and teacher rollouts do not contain explicit \texttt{<think>} traces. Medical specialization uses the native thinking configuration of II-Medical-7B-Preview and Qwen3-8B; both are run with thinking enabled, and the resulting rollouts can contain explicit \texttt{<think>} reasoning. Thus, the two experiments evaluate UCMOPD without and with thinking traces, respectively. The consistent improvements across these settings suggest that UCMOPD is effective under both generation regimes, although we do not conduct a controlled comparison between thinking and no-thinking modes.

\section{Additional Experiments}
\label{app:additional-experiments}

\subsection{Rollout-Budget Control}

Table~\ref{tab:rollout-budget-control} compares vanilla MOPD and UCMOPD under the same eight-response rollout budget.

\begin{table}[h]
\centering
\footnotesize
\setlength{\tabcolsep}{5pt}
\caption{Rollout-budget control with eight sampled responses per prompt.}
\label{tab:rollout-budget-control}
\begin{tabular*}{\linewidth}{@{\extracolsep{\fill}}l*{6}{S[table-format=2.2]}@{}}
\toprule
Method & \multicolumn{1}{c}{Gen. Avg.} & \multicolumn{1}{c}{Vertical Avg.} & \multicolumn{1}{c}{IF-Eval} & \multicolumn{1}{c}{GPQA} & \multicolumn{1}{c}{Zebra} & \multicolumn{1}{c}{Arena-Hard v2} \\
\midrule
MOPD, 8 rollouts & 53.31 & 43.38 & 80.22 & 60.98 & 72.90 & 26.90 \\
UCMOPD, 8 rollouts & \textbf{54.43} & \textbf{45.00} & \textbf{80.96} & \textbf{63.64} & \textbf{76.20} & \textbf{29.35} \\
\bottomrule
\end{tabular*}
\end{table}

\subsection{Golden-Gain Enhancement: Aggregate Diagnostics}

Table~\ref{tab:positive-advantage-density-diagnostics} summarizes the diagnostics visualized across training in Figure~\ref{fig:analysis-diagnostics}. The filter retains $88.6\%$ of exploratory trajectories in this diagnostic run. Retained trajectories have both a larger positive-token fraction and a higher mean positive advantage before Teacher-Endorsement Filtering is applied.

\begin{table}[h]
\centering
\footnotesize
\setlength{\tabcolsep}{6pt}
\caption{Aggregate diagnostics of Golden-Gain Enhancement before Teacher-Endorsement Filtering.}
\label{tab:positive-advantage-density-diagnostics}
\begin{tabular}{l*{2}{S[table-format=1.4]}}
\toprule
\multirow{2}{*}{Subset} & \multicolumn{1}{c}{Positive-token} & \multicolumn{1}{c}{Mean positive} \\
& \multicolumn{1}{c}{fraction $\uparrow$} & \multicolumn{1}{c}{advantage $\uparrow$} \\
\midrule
Kept trajectories & \textbf{0.2168} & \textbf{0.1911} \\
Dropped trajectories & 0.1850 & 0.1158 \\
Kept $-$ dropped & 0.0318 & 0.0752 \\
\bottomrule
\end{tabular}
\end{table}

\subsection{Teacher-Endorsement Filtering: Sample Masking versus Weighting}
\label{app:cll-mode}

The main method uses CLL in sample-mask mode: every nonzero-advantage token update is stochastically kept or dropped according to the direction-consistency probability $w_{\mathrm{cll}}(A_t,y_t)$. We also evaluate a weighting variant that keeps the full token set but multiplies each update by the same probability. Table~\ref{tab:cll-mode-ablation} shows that both variants improve over vanilla MOPD on the general average, but sample masking gives a stronger role-playing average and substantially better GPQA than weighting. Weighting gives a slightly higher general average and slightly better scores on some individual benchmarks in this early CLL-only comparison, but continues to apply reduced updates where masking removes unreliable signals entirely. We therefore use sample masking as the default CLL mechanism in the main method.

\begin{table}[h]
\centering
\footnotesize
\setlength{\tabcolsep}{5pt}
\caption{Implementation comparison for Teacher-Endorsement Filtering without Golden-Gain Enhancement.}
\label{tab:cll-mode-ablation}
\begin{tabular}{l*{6}{S[table-format=2.2]}}
\toprule
Method & \multicolumn{1}{c}{Gen. Avg.} & \multicolumn{1}{c}{Vertical Avg.} & \multicolumn{1}{c}{WritingBench} & \multicolumn{1}{c}{GPQA} & \multicolumn{1}{c}{Zebra} & \multicolumn{1}{c}{LCB v5} \\
\midrule
MOPD & 52.09 & 41.22 & 80.30 & 59.60 & 71.10 & \textbf{35.12} \\
CLL sample mask & 52.17 & \textbf{43.36} & 80.62 & \textbf{62.63} & 72.00 & 34.76 \\
CLL weighting & \textbf{52.34} & 42.92 & \textbf{80.71} & 59.09 & \textbf{72.30} & 31.90 \\
\bottomrule
\end{tabular}
\end{table}

\subsection{Teacher-Endorsement Diagnostics}
\label{app:teacher-endorsement-diagnostics}

We additionally quantify potential direction--endorsement conflict over valid nonzero-advantage tokens in the role-playing diagnostic data. Among negative-advantage tokens, $67.75\%$ coincide with the teacher's top-1 prediction and $76.68\%$ have $s_{\mathrm{cll}}\ge 0.5$, indicating substantial potential over-suppression; among positive-advantage tokens, $12.16\%$ have $s_{\mathrm{cll}}<0.5$, indicating potential over-reinforcement. The top-1 and CLL-threshold rates are descriptive proxies for direction--endorsement conflict rather than ground-truth error labels. The executable filter applies the continuous keep probability $w_{\mathrm{cll}}$ to both positive- and negative-advantage updates rather than using either binary criterion.

The score-based sampling statistics further confirm the expected behavior of the stochastic filter. Table~\ref{tab:cll-expected-realized} reports candidate-token-weighted expected keep rates and the corresponding realized Bernoulli keep rates. Their close agreement shows that the stochastic sample-mask implementation matches its CLL-weighted expectation at the observed token count.

\begin{table}[h]
\centering
\footnotesize
\setlength{\tabcolsep}{7pt}
\caption{Expected and realized keep rates under signed-scope CLL sample masking. Expected rates are candidate-token-weighted means of $w_{\mathrm{cll}}$; realized rates are computed from the sampled Bernoulli masks.}
\label{tab:cll-expected-realized}
\begin{tabular}{lrrr}
\toprule
Candidate subset & Candidate tokens & Expected keep (\%) & Realized keep (\%) \\
\midrule
Negative advantage & 299,307,540 & 9.88 & 9.88 \\
Positive advantage & 115,874,361 & 81.88 & 81.88 \\
All signed candidates & 415,181,901 & 29.97 & 29.97 \\
\bottomrule
\end{tabular}
\end{table}

\begin{figure}[H]
\centering
\includegraphics[width=0.95\linewidth]{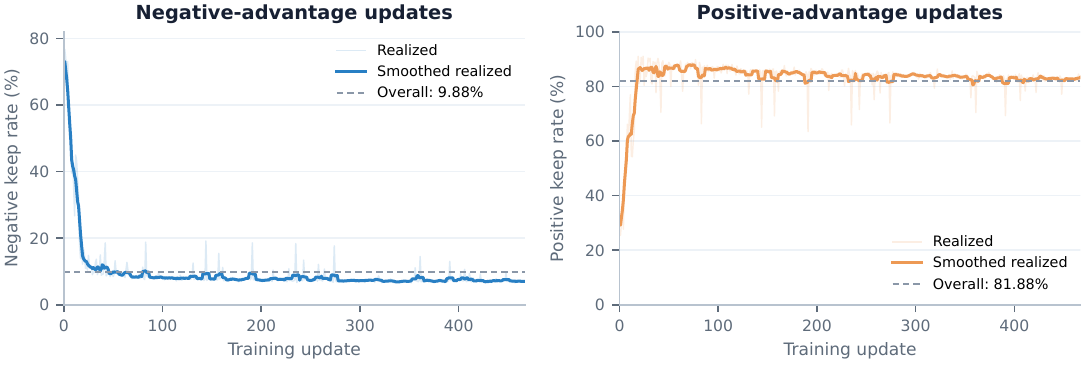}
\caption{Per-update realized CLL retention rates under signed-scope sample masking. The left and right panels show negative- and positive-advantage updates, respectively. Faint traces show the raw rates, solid curves show rolling averages, and dashed lines mark the corresponding overall token-weighted rates reported in Table~\ref{tab:cll-expected-realized}.}
\label{fig:cll-retention-dynamics}
\end{figure}

Figure~\ref{fig:cll-retention-dynamics} directly explains the apparently low negative-branch rate and the asymmetry between the two branches. The per-update realized keep rate is high in the first few updates, then falls and remains near $7$--$8\%$ for most later updates, while the dashed line marks the token-weighted aggregate of $9.88\%$. Because $w^-=1-s_{\mathrm{cll}}$, this decline is consistent with increasing teacher endorsement among sampled negative-advantage tokens: even when the student assigns them slightly higher probability, they are increasingly judged plausible by the teacher and therefore less often retained for suppression. Conversely, $w^+=s_{\mathrm{cll}}$ retains positive-advantage updates more often because their reinforcement directions are more frequently supported by teacher endorsement. This pattern indicates that local student--teacher alignment is improving.

\subsection{Golden-Gain Enhancement: Trajectory-Filtering Criterion}
\label{app:trajectory-filtering-criterion}

Table~\ref{tab:positive-filter} studies the trajectory-level filtering criterion under the same dual-temperature setting. Positive-advantage-density filtering outperforms trajectory-likelihood filtering under both strict and non-strict keep rules, matching our hypothesis that the useful signal is not generic teacher likelihood but the density of teacher-student improvement opportunities. We therefore use the non-strict positive-advantage-density rule in UCMOPD because it provides the strongest role-playing score among the positive-advantage-density variants while retaining a competitive general average.

\begin{table}[t]
\centering
\footnotesize
\setlength{\tabcolsep}{4pt}
\caption{Auxiliary comparison of trajectory filtering criteria under the same dual-temperature setting.}
\label{tab:positive-filter}
\begin{tabular}{l*{4}{S[table-format=2.2]}}
\toprule
Filtering criterion & \multicolumn{1}{c}{Gen. Avg.} & \multicolumn{1}{c}{Vertical Avg.} & \multicolumn{1}{c}{WritingBench} & \multicolumn{1}{c}{ZebraLogic} \\
\midrule
Positive-advantage density ($>$) & 52.87 & 41.46 & 81.26 & 72.10 \\
Likelihood ($>$) & 51.31 & 41.32 & 79.48 & 71.40 \\
Positive-advantage density ($\ge$) & \textbf{53.05} & \textbf{42.66} & \textbf{81.32} & 72.80 \\
Likelihood ($\ge$) & 51.81 & 41.54 & 80.47 & \textbf{74.20} \\
\bottomrule
\end{tabular}
\end{table}

\subsection{Golden-Gain Enhancement: Exploration-Temperature Sensitivity}
\label{app:temperature-ablation}

Table~\ref{tab:temperature-ablation} varies the exploration temperature in Golden-Gain Enhancement. Moderate exploration works best among the tested settings: $T=1.5$ gives the strongest role-playing average and is used as the default in our main configuration. Higher temperatures occasionally improve individual benchmarks such as LiveCodeBench v5 and LiveBench, but are less stable overall, showing that broader exploration does not necessarily yield stronger positive learning signals.

\begin{table}[t]
\centering
\footnotesize
\setlength{\tabcolsep}{5pt}
\caption{Exploration-temperature ablation for positive-advantage-density filtering.}
\label{tab:temperature-ablation}
\begin{tabular}{l*{6}{S[table-format=2.2]}}
\toprule
Temperature & \multicolumn{1}{c}{Gen. Avg.} & \multicolumn{1}{c}{Vertical Avg.} & \multicolumn{1}{c}{IF-Eval} & \multicolumn{1}{c}{Zebra} & \multicolumn{1}{c}{LCB v5} & \multicolumn{1}{c}{LiveBench} \\
\midrule
$1.2$ & 52.30 & 42.09 & 79.11 & \textbf{73.90} & 31.54 & 60.30 \\
$1.5$ & \textbf{53.05} & \textbf{42.66} & \textbf{79.48} & 72.80 & 34.76 & 61.30 \\
$1.8$ & 52.52 & 41.10 & 78.19 & 72.20 & \textbf{39.78} & 61.00 \\
$2.0$ & 52.62 & 42.30 & 79.11 & 73.30 & 33.69 & \textbf{62.30} \\
\bottomrule
\end{tabular}
\end{table}

\section{Properties of the Selection Rules}
\label{app:theory}

We give two simple properties that justify the two mechanisms in UCMOPD. These results are not intended as a global convergence guarantee for neural policy optimization. Instead, they show that Golden-Gain Enhancement provides positive-signal enrichment at the trajectory level, while Teacher-Endorsement Filtering provides entropy-adaptive direction-consistency control at the token level.

\paragraph{Trajectory-Level Positive-Signal Guarantee.}
For each prompt, let $y^a$ be the anchor response and let $\mathcal{E}$ be the set of exploration responses. The filter retains
\[
    \mathcal{E}_{\mathrm{keep}}
    =
    \{y\in\mathcal{E}: r(y)\ge r(y^a)\}.
\]
By construction, every retained exploration response satisfies
\[
    r(y)-r(y^a)\ge 0,\qquad \forall y\in\mathcal{E}_{\mathrm{keep}}.
\]
Consequently, if $\mathcal{E}_{\mathrm{keep}}$ is nonempty, its average positive-advantage-density score is lower bounded by the anchor score:
\[
    \frac{1}{|\mathcal{E}_{\mathrm{keep}}|}
    \sum_{y\in\mathcal{E}_{\mathrm{keep}}} r(y)
    \ge
    r(y^a).
\]
Including the always-retained anchor preserves the same lower bound for the retained candidate set $\{y^a\}\cup\mathcal{E}_{\mathrm{keep}}$. The trajectory filter therefore cannot admit an exploration response whose positive-advantage density is below the prompt-specific anchor threshold.

\paragraph{Entropy-Adaptive Direction--Endorsement Consistency.}
Let $b_T=\exp(-H[p_T])$ denote the entropy-induced typical probability scale of the teacher distribution at a token position. The CLL endorsement score can be rewritten as
\[
    s_{\mathrm{cll}}(x)
    =
    \exp\left(\min\left(\log p_T(x)+H[p_T],0\right)\right)
    =
    \min\left(\frac{p_T(x)}{b_T},1\right).
\]
Accordingly, the unified keep probability becomes
\[
    w_{\mathrm{cll}}(A_t,x)
    =
    \mathbf 1[A_t>0]\min\left(\frac{p_T(x)}{b_T},1\right)
    +
    \mathbf 1[A_t<0]\left[1-\frac{p_T(x)}{b_T}\right]_+,
\]
where $[z]_+=\max(z,0)$. Thus CLL compares the sampled-token probability against an entropy-adaptive baseline rather than a fixed top-$k$ set, and then measures whether the proposed update direction agrees with the resulting endorsement. If $p_T(x)\ge b_T$, endorsement is maximal: increasing the token is fully consistent, while decreasing it is inconsistent. If $p_T(x)<b_T$, endorsement decreases smoothly as $p_T(x)/b_T$, continuously transferring retention probability from reinforcement to suppression.

This form also gives the desired entropy dependence. For fixed $p_T(x)$ below the typical scale, $s_{\mathrm{cll}}(x)=p_T(x)\exp(H[p_T])$ is non-decreasing in teacher entropy. A broader teacher distribution therefore recognizes more tokens as plausible, while a confident teacher assigns lower relative endorsement to the same low-probability token. The consistency rule automatically turns this single change in endorsement into opposite retention tendencies for probability-increasing and probability-decreasing updates.

\end{document}

%% file: math_commands.tex
\usepackage{amsmath,amsfonts,bm}

\def\eqref#1{equation~\ref{#1}}

\def\1{\bm{1}}

\DeclareMathAlphabet{\mathsfit}{\encodingdefault}{\sfdefault}{m}{sl}
\SetMathAlphabet{\mathsfit}{bold}{\encodingdefault}{\sfdefault}{bx}{n}



%% file: references.bib
@misc{ii_medical_7b_preview,
  title={II-Medical-7B-Preview: Medical Reasoning Model}, 
  author={Intelligent Internet},
  year={2025},
  url={https://huggingface.co/Intelligent-Internet/II-Medical-7B-Preview}
}

@article{cox1961tests,
  title={Tests of separate families of hypotheses},
  author={Cox, David Roxbee},
  year={1961}
}

@article{zhang2026onpolicysft,
  title={Towards On-Policy SFT: Distribution Discriminant Theory and Its Applications in LLM Training},
  author={Zhang, Miaosen and Liu, Yishan and Lin, Shuxia and Yang, Xu and Dai, Qi and Luo, Chong and Jiang, Weihao and Hou, Peng and Zeng, Anxiang and Geng, Xin and Guo, Baining},
  journal={arXiv preprint arXiv:2602.12222},
  year={2026},
  url={https://arxiv.org/abs/2602.12222}
}

@article{xiao2026mimo,
  title={{MiMo-V2-Flash} Technical Report},
  author={Xiao, Bangjun and Xia, Bingquan and Yang, Bo and Gao, Bofei and Shen, Bowen and He, Chenhong and Lou, Chiheng and Luo, Fuli and Wang, Gang and others},
  journal={arXiv preprint arXiv:2601.02780},
  year={2026},
  url={https://arxiv.org/abs/2601.02780}
}

@article{dou2026baichuan,
  title={{Baichuan-M3}: Modeling Clinical Inquiry for Reliable Medical Decision-Making},
  author={Dou, Chengfeng and Yang, Fan and Li, Fei and Jia, Jiyuan and Ju, Qiang and Wang, Shuai and Li, Tianpeng and Zeng, Xiangrong and Zhou, Yijie and Zhang, Hongda and others},
  journal={arXiv preprint arXiv:2602.06570},
  year={2026},
  url={https://arxiv.org/abs/2602.06570}
}

@misc{wang2025coser,
  title={CoSER: A Comprehensive Literary Dataset and Framework for Training and Evaluating LLM Role-Playing and Persona Simulation}, 
  author={Xintao Wang and Heng Wang and Yifei Zhang and Xinfeng Yuan and Rui Xu and Jen-tse Huang and Siyu Yuan and Haoran Guo and Jiangjie Chen and Shuchang Zhou and Wei Wang and Yanghua Xiao},
  year={2026},
  eprint={2502.09082},
  archivePrefix={arXiv},
  primaryClass={cs.CL},
  url={https://arxiv.org/abs/2502.09082}, 
}

@article{jin2020disease,
  title={What Disease Does This Patient Have? A Large-scale Open Domain Question Answering Dataset from Medical Exams},
  author={Jin, Di and Pan, Eileen and Oufattole, Nassim and Weng, Wei-Hung and Fang, Hanyi and Szolovits, Peter},
  journal={Applied Sciences},
  volume={11},
  number={14},
  pages={6421},
  year={2021},
  url={https://www.mdpi.com/2076-3417/11/14/6421}
}

@article{gema2024done,
  title={Are We Done with {MMLU}?},
  author={Gema, Aryo Pradipta and Leang, Joshua Ong Jun and Hong, Giwon and Devoto, Alessio and Mancino, Alberto Carlo Maria and Saxena, Rohit and He, Xuanli and Zhao, Yu and Du, Xiaotang and Madani, Mohammad Reza Ghasemi and Barale, Claire and McHardy, Robert and Harris, Joshua and Kaddour, Jean and van Krieken, Emile and Minervini, Pasquale},
  journal={arXiv preprint arXiv:2406.04127},
  year={2024},
  url={https://arxiv.org/abs/2406.04127}
}

@article{wang2024mmlu,
  title={{MMLU-Pro}: A More Robust and Challenging Multi-Task Language Understanding Benchmark},
  author={Wang, Yubo and Ma, Xueguang and Zhang, Ge and Ni, Yuntian and others},
  journal={arXiv preprint arXiv:2406.01574},
  year={2024},
  url={https://arxiv.org/abs/2406.01574}
}

@article{rein2023gpqa,
  title={{GPQA}: A Graduate-Level Google-Proof Q\&A Benchmark},
  author={Rein, David and Hou, Betty Li and Stickland, Asa Cooper and Petty, Jackson and Pang, Richard Yuanzhe and Dirani, Julien and Michael, Julian and Bowman, Samuel R.},
  journal={arXiv preprint arXiv:2311.12022},
  year={2023},
  url={https://arxiv.org/abs/2311.12022}
}

@article{balunovic2025matharena,
  title={{MathArena}: Evaluating {LLM}s on Uncontaminated Math Competitions},
  author={Balunovic, Mislav and Dekoninck, Jasper and Petrov, Ivo and Jovanovic, Nikola and Vechev, Martin},
  journal={arXiv preprint arXiv:2505.23281},
  year={2025},
  url={https://arxiv.org/abs/2505.23281}
}

@article{lin2025zebralogic,
  title={{ZebraLogic}: On the Scaling Limits of {LLM}s for Logical Reasoning},
  author={Lin, Bill Yuchen and Le Bras, Ronan and Richardson, Kyle and Sabharwal, Ashish and Poovendran, Radha and Clark, Peter and Choi, Yejin},
  journal={arXiv preprint arXiv:2502.01100},
  year={2025},
  url={https://arxiv.org/abs/2502.01100}
}

@article{jain2024livecodebench,
  title={{LiveCodeBench}: Holistic and Contamination Free Evaluation of Large Language Models for Code},
  author={Jain, Naman and Han, King and Gu, Alex and Li, Wen-Ding and Yan, Fanjia and Zhang, Tianjun and Wang, Sida and Solar-Lezama, Armando and Sen, Koushik and Stoica, Ion},
  journal={arXiv preprint arXiv:2403.07974},
  year={2024},
  url={https://arxiv.org/abs/2403.07974}
}

@article{zhou2023ifeval,
  title={Instruction-Following Evaluation for Large Language Models},
  author={Zhou, Jeffrey and Lu, Tianjian and Mishra, Swaroop and Brahma, Siddhartha and Basu, Sujoy and Luan, Yi and Zhou, Denny and Hou, Le},
  journal={arXiv preprint arXiv:2311.07911},
  year={2023},
  url={https://arxiv.org/abs/2311.07911}
}

@inproceedings{pyatkin2025generalizing,
  title={Generalizing Verifiable Instruction Following},
  author={Pyatkin, Valentina and Malik, Saumya and Graf, Victoria and Ivison, Hamish and Huang, Shengyi and Dasigi, Pradeep and Lambert, Nathan and Hajishirzi, Hanna},
  booktitle={Advances in Neural Information Processing Systems 38},
  pages={54870--54896},
  year={2025},
  doi={10.52202/085713-1645},
  url={https://arxiv.org/abs/2507.02833}
}

@article{wu2025writingbench,
  title={{WritingBench}: A Comprehensive Benchmark for Generative Writing},
  author={Wu, Yuning and Mei, Jiahao and Yan, Ming and Li, Chenliang and Lai, Shaopeng and Ren, Yuran and Wang, Zijia and Zhang, Ji and Wu, Mengyue and Jin, Qin and Huang, Fei},
  journal={arXiv preprint arXiv:2503.05244},
  year={2025},
  url={https://arxiv.org/abs/2503.05244}
}

@article{li2024arenahard,
  title={From Crowdsourced Data to High-Quality Benchmarks: {Arena-Hard} and {BenchBuilder} Pipeline},
  author={Li, Tianle and Chiang, Wei-Lin and Frick, Evan and Dunlap, Lisa and Wu, Tianhao and Zhu, Banghua and Gonzalez, Joseph E. and Stoica, Ion},
  journal={arXiv preprint arXiv:2406.11939},
  year={2024},
  url={https://arxiv.org/abs/2406.11939}
}

@inproceedings{white2025livebench,
  title={{LiveBench}: A Challenging, Contamination-Limited {LLM} Benchmark},
  author={White, Colin and Dooley, Samuel and Roberts, Manley and Pal, Arka and Feuer, Benjamin and Jain, Siddhartha and Shwartz-Ziv, Ravid and Jain, Neel and Saifullah, Khalid and Dey, Sreemanti and Shubh-Agrawal and Sandha, Sandeep Singh and Naidu, Siddartha and Hegde, Chinmay and LeCun, Yann and Goldstein, Tom and Neiswanger, Willie and Goldblum, Micah},
  booktitle={International Conference on Learning Representations},
  year={2025},
  url={https://arxiv.org/abs/2406.19314}
}

@article{zuo2025medxpertqa,
  title={{MedXpertQA}: Benchmarking Expert-Level Medical Reasoning and Understanding},
  author={Zuo, Yuxin and Qu, Shang and Li, Yifei and Chen, Zhangren and Zhu, Xuekai and Hua, Ermo and Zhang, Kaiyan and Ding, Ning and Zhou, Bowen},
  journal={arXiv preprint arXiv:2501.18362},
  year={2025},
  url={https://arxiv.org/abs/2501.18362}
}

@inproceedings{jin2019pubmedqa,
  title={{PubMedQA}: A Dataset for Biomedical Research Question Answering},
  author={Jin, Qiao and Dhingra, Bhuwan and Liu, Zhengping and Cohen, William W. and Lu, Xinghua},
  booktitle={Proceedings of the 2019 Conference on Empirical Methods in Natural Language Processing and the 9th International Joint Conference on Natural Language Processing},
  pages={2567--2577},
  year={2019},
  url={https://aclanthology.org/D19-1259/}
}

@article{hinton2015distilling,
  title={Distilling the Knowledge in a Neural Network},
  author={Hinton, Geoffrey and Vinyals, Oriol and Dean, Jeff},
  journal={arXiv preprint arXiv:1503.02531},
  year={2015},
  url={https://arxiv.org/abs/1503.02531}
}

@inproceedings{kim2016sequence,
  title={Sequence-Level Knowledge Distillation},
  author={Kim, Yoon and Rush, Alexander M.},
  booktitle={Proceedings of the 2016 Conference on Empirical Methods in Natural Language Processing},
  pages={1317--1327},
  year={2016},
  url={https://arxiv.org/abs/1606.07947}
}

@inproceedings{ross2011dagger,
  title={A Reduction of Imitation Learning and Structured Prediction to No-Regret Online Learning},
  author={Ross, Stephane and Gordon, Geoffrey J. and Bagnell, J. Andrew},
  booktitle={Proceedings of the Fourteenth International Conference on Artificial Intelligence and Statistics},
  pages={627--635},
  year={2011},
  url={https://arxiv.org/abs/1011.0686}
}

@inproceedings{gkd,
  title={On-Policy Distillation of Language Models: Learning from Self-Generated Mistakes},
  author={Agarwal, Rishabh and Vieillard, Nino and Zhou, Yongchao and Stanczyk, Piotr and Ramos, Sabela and Geist, Matthieu and Bachem, Olivier},
  booktitle={International Conference on Learning Representations},
  year={2024},
  url={https://arxiv.org/abs/2306.13649}
}

@inproceedings{minillm,
  title={{MiniLLM}: On-Policy Distillation of Large Language Models},
  author={Gu, Yuxian and Dong, Li and Wei, Furu and Huang, Minlie},
  booktitle={International Conference on Learning Representations},
  year={2024},
  url={https://arxiv.org/abs/2306.08543}
}

@inproceedings{distillm,
  title={{DistiLLM}: Towards Streamlined Distillation for Large Language Models},
  author={Ko, Jongwoo and Kim, Sungnyun and Chen, Tianyi and Yun, Se-Young},
  booktitle={Proceedings of the 41st International Conference on Machine Learning},
  year={2024},
  url={https://arxiv.org/abs/2402.03898}
}

@inproceedings{distillm2,
  title={{DistiLLM-2}: A Contrastive Approach Boosts the Distillation of {LLM}s},
  author={Ko, Jongwoo and Chen, Tianyi and Kim, Sungnyun and Ding, Tianyu and Liang, Luming and Zharkov, Ilya and Yun, Se-Young},
  booktitle={Proceedings of the 42nd International Conference on Machine Learning},
  year={2025},
  url={https://arxiv.org/abs/2503.07067}
}

@article{reopold,
  title={Scaling Reasoning Efficiently via Relaxed On-Policy Distillation},
  author={Ko, Jongwoo and Abdali, Sara and Kim, Young Jin and Chen, Tianyi and Cameron, Pashmina},
  journal={arXiv preprint arXiv:2603.11137},
  year={2026},
  url={https://arxiv.org/abs/2603.11137}
}

@article{selectkd,
  title={{SelecTKD}: Selective Token-Weighted Knowledge Distillation for {LLM}s},
  author={Huang, Haiduo and Song, Jiangcheng and Zhang, Yadong and Ren, Pengju},
  journal={arXiv preprint arXiv:2510.24021},
  year={2025},
  url={https://arxiv.org/abs/2510.24021}
}

@article{scope_opd,
  title={{SCOPE}: Signal-Calibrated On-Policy Distillation Enhancement with Dual-Path Adaptive Weighting},
  author={Zheng, Binbin and Ma, Xing and Liang, Yiheng and Ruan, Jingqing and Fu, Xiaoliang and Lin, Kepeng and Zhu, Benchang and Zeng, Ke and Cai, Xunliang},
  journal={arXiv preprint arXiv:2604.10688},
  year={2026},
  url={https://arxiv.org/abs/2604.10688}
}

@article{eopd_jin2026,
  title={Entropy-aware on-policy distillation of language models},
  author={Jin, Woogyeol and Min, Taywon and Yang, Yongjin and Kadhe, Swanand Ravindra and Zhou, Yi and Wei, Dennis and Baracaldo, Nathalie and Lee, Kimin},
  journal={arXiv preprint arXiv:2603.07079},
  year={2026},
  url={https://arxiv.org/abs/2603.07079}
}

@article{yang2026gopd,
  title={Learning beyond Teacher: Generalized On-Policy Distillation with Reward Extrapolation},
  author={Yang, Wenkai and Liu, Weijie and Xie, Ruobing and Yang, Kai and Yang, Saiyong and Lin, Yankai},
  journal={arXiv preprint arXiv:2602.12125},
  year={2026},
  url={https://arxiv.org/abs/2602.12125}
}

@article{prefix_opd,
  title={Fast and Effective On-policy Distillation from Reasoning Prefixes},
  author={Zhang, Dongxu and Yang, Zhichao and Janghorbani, Sepehr and Han, Jun and Ressler, Andrew and Qian, Qian and Lyng, Gregory D. and Batra, Sanjit Singh and Tillman, Robert E.},
  journal={arXiv preprint arXiv:2602.15260},
  year={2026},
  url={https://arxiv.org/abs/2602.15260}
}

@article{paced,
  title={{PACED}: Distillation and On-Policy Self-Distillation at the Frontier of Student Competence},
  author={Xu, Yuanda and Sang, Hejian and Zhou, Zhengze and He, Ran and Wang, Zhipeng},
  journal={arXiv preprint arXiv:2603.11178},
  year={2026},
  url={https://arxiv.org/abs/2603.11178}
}

@inproceedings{skd,
  title={Speculative Knowledge Distillation: Bridging the Teacher-Student Gap Through Interleaved Sampling},
  author={Xu, Wenda and Han, Rujun and Wang, Zifeng and Le, Long T. and Madeka, Dhruv and Li, Lei and Wang, William Yang and Agarwal, Rishabh and Lee, Chen-Yu and Pfister, Tomas},
  booktitle={International Conference on Learning Representations},
  year={2025},
  url={https://arxiv.org/abs/2410.11325}
}

@inproceedings{adakd,
  title={{LLM}-Oriented Token-Adaptive Knowledge Distillation},
  author={Xie, Xurong and Xue, Zhucun and Wu, Jiafu and Li, Jian and Wang, Yabiao and Hu, Xiaobin and Liu, Yong and Zhang, Jiangning},
  booktitle={Proceedings of the AAAI Conference on Artificial Intelligence},
  year={2026},
  note={Also available as arXiv:2510.11615},
  url={https://arxiv.org/abs/2510.11615}
}

@article{token_selective_dual_kd,
  title={Explain in Your Own Words: Improving Reasoning via Token-Selective Dual Knowledge Distillation},
  author={Kim, Minsang and Baek, Seung Jun},
  journal={arXiv preprint arXiv:2603.13260},
  year={2026},
  url={https://arxiv.org/abs/2603.13260}
}

@article{opd_survey_song2026,
  title={A Survey of On-Policy Distillation for Large Language Models},
  author={Song, Mingyang and Zheng, Mao},
  journal={arXiv preprint arXiv:2604.00626},
  year={2026},
  url={https://arxiv.org/abs/2604.00626}
}

@inproceedings{you2017learning,
  title={Learning from multiple teacher networks},
  author={You, Shan and Xu, Chang and Xu, Chao and Tao, Dacheng},
  booktitle={Proceedings of the 23rd ACM SIGKDD international conference on knowledge discovery and data mining},
  pages={1285--1294},
  year={2017}
}

@inproceedings{fukuda2017efficient,
  title={Efficient knowledge distillation from an ensemble of teachers.},
  author={Fukuda, Takashi and Suzuki, Masayuki and Kurata, Gakuto and Thomas, Samuel and Cui, Jia and Ramabhadran, Bhuvana},
  booktitle={Interspeech},
  pages={3697--3701},
  year={2017}
}

@article{nemotron_cascade2,
  title={{Nemotron-Cascade 2}: Post-Training {LLM}s with Cascade {RL} and Multi-Domain On-Policy Distillation},
  author={Yang, Zhuolin and Liu, Zihan and Chen, Yang and Dai, Wenliang and Wang, Boxin and Lin, Sheng-Chieh and Lee, Chankyu and Chen, Yangyi and Jiang, Dongfu and He, Jiafan and others},
  journal={arXiv preprint arXiv:2603.19220},
  year={2026},
  url={https://arxiv.org/abs/2603.19220}
}

@article{deepseek_v4_report,
  title={Deepseek-v4: Towards highly efficient million-token context intelligence},
  author={Xu, Anyi and Lin, Bangcai and Xue, Bing and Wang, Bingxuan and Xu, Bingzheng and Wu, Bochao and Zhang, Bowei and Lin, Chaofan and Dong, Chen and Ling, Chenchen and others},
  journal={arXiv preprint arXiv:2606.19348},
  year={2026}
}

@article{camopd_chen2026,
  title={Counteraction-Aware Multi-Teacher On-Policy Distillation for General Capability Recovery with Domain Preservation},
  author={Chen, Tianlei and Ou, Jiao and Liu, Ziyuan and Tang, Ruiming and Liang, Jian and Li, Han},
  journal={arXiv preprint arXiv:2605.27115},
  year={2026},
  url={https://arxiv.org/abs/2605.27115}
}

@article{mccloskey1989catastrophic,
  title={Catastrophic Interference in Connectionist Networks: The Sequential Learning Problem},
  author={McCloskey, Michael and Cohen, Neal J.},
  journal={Psychology of Learning and Motivation},
  volume={24},
  pages={109--165},
  year={1989}
}

@article{kirkpatrick2017overcoming,
  title={Overcoming Catastrophic Forgetting in Neural Networks},
  author={Kirkpatrick, James and Pascanu, Razvan and Rabinowitz, Neil and Veness, Joel and Desjardins, Guillaume and Rusu, Andrei A. and Milan, Kieran and Quan, John and Ramalho, Tiago and Grabska-Barwinska, Agnieszka and others},
  journal={Proceedings of the National Academy of Sciences},
  volume={114},
  number={13},
  pages={3521--3526},
  year={2017},
  url={https://arxiv.org/abs/1612.00796}
}

@article{lopezpaz2017gem,
  title={Gradient episodic memory for continual learning},
  author={Lopez-Paz, David and Ranzato, Marc'Aurelio},
  journal={Advances in neural information processing systems},
  volume={30},
  year={2017}
}

@article{rolnick2019experience,
  title={Experience replay for continual learning},
  author={Rolnick, David and Ahuja, Arun and Schwarz, Jonathan and Lillicrap, Timothy and Wayne, Gregory},
  journal={Advances in neural information processing systems},
  volume={32},
  year={2019}
}

@article{lesort2020continual,
  title={Continual learning: Tackling catastrophic forgetting in deep neural networks with replay processes},
  author={Lesort, Timoth{\'e}e},
  journal={arXiv preprint arXiv:2007.00487},
  year={2020}
}

@article{luo2023empirical,
  title={An Empirical Study of Catastrophic Forgetting in Large Language Models During Continual Fine-tuning},
  author={Luo, Yun and Yang, Zhen and Meng, Fandong and Li, Yafu and Zhou, Jie and Zhang, Yue},
  journal={arXiv preprint arXiv:2308.08747},
  year={2023},
  url={https://arxiv.org/abs/2308.08747}
}

@inproceedings{dong2024abilities,
  title={How Abilities in Large Language Models are Affected by Supervised Fine-tuning Data Composition},
  author={Dong, Guanting and Yuan, Hongyi and Lu, Keming and Li, Chengpeng and Xue, Mingfeng and Liu, Dayiheng and Wang, Wei and Yuan, Zheng and Zhou, Chang and Zhou, Jingren},
  booktitle={Proceedings of the 62nd Annual Meeting of the Association for Computational Linguistics},
  year={2024},
  url={https://arxiv.org/abs/2310.05492}
}

@inproceedings{bethune2025scaling,
  title={Scaling Laws for Forgetting during Finetuning with Pretraining Data Injection},
  author={Bethune, Louis and Grangier, David and Busbridge, Dan and Gualdoni, Eleonora and Cuturi, Marco and Ablin, Pierre},
  booktitle={Proceedings of the 42nd International Conference on Machine Learning},
  year={2025},
  url={https://arxiv.org/abs/2502.06042}
}

@inproceedings{huang2024self,
  title={Mitigating Catastrophic Forgetting in Large Language Models with Self-Synthesized Rehearsal},
  author={Huang, Jianheng and Cui, Leyang and Wang, Ante and Yang, Chengyi and Liao, Xinting and Song, Linfeng and Yao, Junfeng and Su, Jinsong},
  booktitle={Proceedings of the 62nd Annual Meeting of the Association for Computational Linguistics},
  pages={1416--1428},
  year={2024},
  url={https://aclanthology.org/2024.acl-long.77/}
}

@article{chen2024criticaldomains,
  title={A Survey on Large Language Models for Critical Societal Domains: Finance, Healthcare, and Law},
  author={Chen, Zhiyu Zoey and Ma, Jing and Zhang, Xinlu and Hao, Nan and Yan, An and Nourbakhsh, Armineh and Yang, Xianjun and McAuley, Julian and Petzold, Linda and Wang, William Yang},
  journal={arXiv preprint arXiv:2405.01769},
  year={2024},
  url={https://arxiv.org/abs/2405.01769}
}

@article{lu2024domainadaptation,
  title={Fine-tuning Large Language Models for Domain Adaptation: Exploration of Training Strategies, Scaling, Model Merging and Synergistic Capabilities},
  author={Lu, Wei and Luu, Rachel K. and Buehler, Markus J.},
  journal={arXiv preprint arXiv:2409.03444},
  year={2024},
  url={https://arxiv.org/abs/2409.03444}
}

@article{yang2025specializedllm,
  title={Survey of Specialized Large Language Model},
  author={Yang, Chenghan and Zhao, Ruiyu and Liu, Yang and Jiang, Ling},
  journal={arXiv preprint arXiv:2508.19667},
  year={2025},
  url={https://arxiv.org/abs/2508.19667}
}

@inproceedings{liu2024gci,
  title={More than catastrophic forgetting: Integrating general capabilities for domain-specific llms},
  author={Liu, Chengyuan and Kang, Yangyang and Wang, Shihang and Qing, Lizhi and Zhao, Fubang and Wu, Chao and Sun, Changlong and Kuang, Kun and Wu, Fei},
  booktitle={Proceedings of the 2024 Conference on Empirical Methods in Natural Language Processing},
  pages={7531--7548},
  year={2024}
}

@misc{yang2025qwen3technicalreport,
      title={Qwen3 Technical Report}, 
      author={An Yang and Anfeng Li and Baosong Yang and Beichen Zhang and Binyuan Hui and Bo Zheng and Bowen Yu and Chang Gao and Chengen Huang and Chenxu Lv and Chujie Zheng and Dayiheng Liu and Fan Zhou and Fei Huang and Feng Hu and Hao Ge and Haoran Wei and Huan Lin and Jialong Tang and Jian Yang and Jianhong Tu and Jianwei Zhang and Jianxin Yang and Jiaxi Yang and Jing Zhou and Jingren Zhou and Junyang Lin and Kai Dang and Keqin Bao and Kexin Yang and Le Yu and Lianghao Deng and Mei Li and Mingfeng Xue and Mingze Li and Pei Zhang and Peng Wang and Qin Zhu and Rui Men and Ruize Gao and Shixuan Liu and Shuang Luo and Tianhao Li and Tianyi Tang and Wenbiao Yin and Xingzhang Ren and Xinyu Wang and Xinyu Zhang and Xuancheng Ren and Yang Fan and Yang Su and Yichang Zhang and Yinger Zhang and Yu Wan and Yuqiong Liu and Zekun Wang and Zeyu Cui and Zhenru Zhang and Zhipeng Zhou and Zihan Qiu},
      year={2025},
      eprint={2505.09388},
      archivePrefix={arXiv},
      primaryClass={cs.CL},
      url={https://arxiv.org/abs/2505.09388}, 
}

@article{bercovich2025llamanemotron,
  title={{Llama-Nemotron}: Efficient Reasoning Models},
  author={Bercovich, Akhiad and Levy, Itay and Golan, Izik and Dabbah, Mohammad and El-Yaniv, Ran and others},
  journal={arXiv preprint arXiv:2505.00949},
  year={2025},
  url={https://arxiv.org/abs/2505.00949}
}

@inproceedings{pal2022medmcqa,
  title={{MedMCQA}: A Large-scale Multi-Subject Multi-Choice Dataset for Medical Domain Question Answering},
  author={Pal, Ankit and Umapathi, Logesh Kumar and Sankarasubbu, Malaikannan},
  booktitle={Proceedings of the Conference on Health, Inference, and Learning},
  volume={174},
  pages={248--260},
  year={2022},
  url={https://proceedings.mlr.press/v174/pal22a.html}
}

@article{wu2025medreason,
  title={{MedReason}: Eliciting Factual Medical Reasoning Steps in {LLM}s via Knowledge Graphs},
  author={Wu, Juncheng and Deng, Wenlong and Li, Xingxuan and Liu, Sheng and Mi, Taomian and Peng, Yifan and Xu, Ziyang and Liu, Yi and Cho, Hyunjin and Choi, Chang-In and others},
  journal={arXiv preprint arXiv:2504.00993},
  year={2025},
  url={https://arxiv.org/abs/2504.00993}
}

@article{chen2024huatuogpto1,
  title={{HuatuoGPT-o1}: Towards Medical Complex Reasoning with {LLM}s},
  author={Chen, Junying and Cai, Zhenyang and Ji, Ke and Wang, Xidong and Liu, Wanlong and Wang, Rongsheng and Hou, Jianye and Wang, Benyou},
  journal={arXiv preprint arXiv:2412.18925},
  year={2024},
  url={https://arxiv.org/abs/2412.18925}
}

@article{huang2025m1,
  title={{m1}: Unleash the Potential of Test-Time Scaling for Medical Reasoning with Large Language Models},
  author={Huang, Xiaoke and Wu, Juncheng and Liu, Hui and Tang, Xianfeng and Zhou, Yuyin},
  journal={arXiv preprint arXiv:2504.00869},
  year={2025},
  url={https://arxiv.org/abs/2504.00869}
}
